\documentclass[journal]{IEEEtran}
\usepackage{cite}

\ifCLASSINFOpdf
\else
\fi

\usepackage{amsmath,amssymb,amsthm,bm,bbm,graphicx,color}
\usepackage{url}
\usepackage[colorlinks, linkcolor=blue, anchorcolor=blue, citecolor=blue]{hyperref}
\usepackage{subfigure}
\usepackage{pifont}
\newcommand{\cmark}{\ding{51}}%
\newcommand{\xmark}{\ding{55}}%

\begin{document}
%
\title{Scalable Gaussian Process Classification with Additive Noise for Various Likelihoods}
%
%
%

\author{
	Haitao~Liu,
	Yew-Soon~Ong,~\IEEEmembership{Fellow,~IEEE,},
	Ziwei Yu,	
	Jianfei~Cai,~\IEEEmembership{Senior~Member,~IEEE},
	and~Xiaobo Shen 
\thanks{Haitao Liu is with the Rolls-Royce@NTU Corporate Lab, Nanyang Technological University, Singapore, 637460. E-mail: htliu@ntu.edu.sg.}
\thanks{Yew-Soon Ong and Jianfei Cai are with School of Computer Science and Engineering, Nanyang Technological University, Singapore, 639798. E-mail: oyewsoon@gmail.com, asjfcai@ntu.edu.sg.}
\thanks{Ziwei Yu is with School of Computing, National University of Singapore, Singapore, 117417. Email: yuziwei@u.nus.edu.}
\thanks{Xiaobo Shen is with School of Computer Science and Engineering, Nanjing University of Science and Technology, China. Email: xbshen@njust.edu.cn.}
}

%
%

\markboth{}%
{Shell \MakeLowercase{\textit{et al.}}: Bare Demo of IEEEtran.cls for IEEE Journals}
%



\maketitle

\begin{abstract}
Gaussian process classification (GPC) provides a flexible and powerful statistical framework describing joint distributions over function space. Conventional GPCs however suffer from (i) poor scalability for big data due to the full kernel matrix, and (ii) intractable inference due to the non-Gaussian likelihoods. Hence, various scalable GPCs have been proposed through (i) the sparse approximation built upon a small inducing set to reduce the time complexity; and (ii) the approximate inference to derive analytical evidence lower bound (ELBO). However, these scalable GPCs equipped with analytical ELBO are limited to specific likelihoods or additional assumptions. In this work, we present a \textit{unifying} framework which accommodates scalable GPCs using various likelihoods. Analogous to GP regression (GPR), we introduce additive noises to augment the probability space for (i) the GPCs with \textit{step}, (multinomial) \textit{probit} and \textit{logit} likelihoods via the internal variables; and \textit{particularly}, (ii) the GPC using \textit{softmax} likelihood via the noise variables themselves. This leads to unified scalable GPCs with analytical ELBO by using variational inference. Empirically, our GPCs showcase better results than state-of-the-art scalable GPCs for extensive binary/multi-class classification tasks with up to two million data points.
\end{abstract}


%
\IEEEpeerreviewmaketitle

\section{Introduction}
%
%
%
%
\IEEEPARstart{A}{s} a non-parametric Bayesian model which is explainable and provides confidence in predictions, Gaussian process (GP) has been widely investigated and used in various scenarios, e.g., regression and classification~\cite{rasmussen2006gaussian}, active learning~\cite{settles1994active}, unsupervised learning~\cite{lawrence2005probabilistic}, and multi-task learning~\cite{alvarez2012kernels, liu2018remarks}. The central task in GP is to infer the latent function $f$, which follows a Gaussian process $\mathcal{GP}(0, k(.))$ where the kernel $k(.)$ describes the covariance among inputs, from $n$ observations $\mathbf{X}=(\mathbf{x}_1, \cdots, \mathbf{x}_n)^{\mathsf{T}}$ with labels $\mathbf{y}=(y_1, \cdots, y_n)^{\mathsf{T}}$. The inference can be performed through the type-II maximum likelihood which maximizes over the model evidence $p(\mathbf{y})$.

We herein focus on GP classification (GPC)~\cite{kim2006bayesian, wang2013spectrum} with discrete class labels. It is more challenging than the GP regression (GPR). Specifically, current GPC paradigms are facing two main challenges. The first is the poor scalability to tackle massive datasets. The inversion and determinant of the kernel matrix $\mathbf{K}_{nn} = k(\mathbf{X}, \mathbf{X}) \in R^{n \times n}$ incur $\mathcal{O}(n^3)$ time complexity for inference. This cubic complexity severely limits the applicability of GP, especially in the era of big data. It becomes more serious for multi-class classification, since we need to infer $C$ latent functions for $C$ classes. The second is the intractable inference for the posterior $p(\mathbf{f}|\mathbf{y})$ where $\mathbf{f}$ is the latent function values at data points. Due to the commonly used non-Gaussian likelihoods $p(\mathbf{y}|\mathbf{f})$, e.g., the \textit{step} likelihood, the (multinomial) \textit{probit/logit} likelihoods and the \textit{softmax} likelihood, the Bayesian rule $p(\mathbf{f}|\mathbf{y}) \propto p(\mathbf{y}|\mathbf{f}) p(\mathbf{f})$ incurs however intractable inference. 

Inspired by the success of scalable GPR in recent years~\cite{quinonero2005unifying, titsias2009variational, hensman2013gaussian}, alternatively, we could address the two major issues in GPC by regarding the classification with discrete labels as a regression task. For example, we could either directly treat GPC as GPR, like~\cite{frohlich2013large}; or we interpret the class labels as the outputs of a Dirichlet distribution to encourage the GPR-like inference~\cite{milios2018dirichlet}.

A more principled alternative is adopting the GPC framework to handle binary/multi-class cases. To address the inference issue, various approximate inference algorithms, e.g., laplace approximation (LA), expectation propagation (EP) and variational inference (VI), have been developed, the core of which is approximating the non-Gaussian $p(\mathbf{f}|\mathbf{y})$ with a tractable Gaussian $q(\mathbf{f})$~\cite{nickisch2008approximations}.

As for the scalability issue, it has been extensively exploited in the regime of GPR~\cite{liu2018gaussian}. Particularly, the sparse approximations~\cite{snelson2006sparse, titsias2009variational, hensman2013gaussian, wilson2015kernel} seek to distill the entire training data through a global inducing set $\{\mathbf{X}_m, \mathbf{u}\}$ comprising $m$ ($m \ll n$) points, thus reducing the time complexity from $\mathcal{O}(n^3)$ to $\mathcal{O}(nm^2)$. This is achieved either by modifying the joint prior as $p(\mathbf{f}, f_*) \approx q(\mathbf{f}, f_*)$ where $f_*$ is the latent function value at the test point $\mathbf{x}_*$~\cite{quinonero2005unifying}; or through directly approximating the posterior $p(\mathbf{f}|\mathbf{y}) \approx q(\mathbf{f})$~\cite{titsias2009variational}. The time complexity can be further reduced to $\mathcal{O}(m^3)$ by recognizing an evidence lower bound (ELBO) for $\log p(\mathbf{y})$. The ELBO factorizes over data points~\cite{hensman2013gaussian, hoang2015unifying, peng2017asynchronous}, thus allowing efficient stochastic variational inference~\cite{hoffman2013stochastic}. Moreover, by exploiting the specific structures, e.g., the Kronecker and Toeplitz structures, in the inducing set, the time complexity can be dramatically reduced to $\mathcal{O}(n)$~\cite{wilson2015kernel, pleiss2018constant, gardner2018product}.

Hence, scalable GPCs could inherit the sparse framework from GPR, with the difficulty being that the model evidence or ELBO should be carefully built up in order to overcome the intractable Gaussian integrals. To this end, the fully independent training conditional (FITC) assumption $p(\mathbf{f}|\mathbf{u}) = \prod_{i=1}^n p(f_i|\mathbf{u})$~\cite{snelson2006sparse} is employed to build scalable binary GPC~\cite{naish2008generalized}.\footnote{For GPR, this assumption severally underestimates the noise variance and worsens the prediction mean~\cite{bauer2016understanding}.} The scalability has been further improved for binary/multi-class classification through the stochastic variants~\cite{hernandez2016scalable, villacampa2017scalable}, which derive a closed-form ELBO factorized over data points, thus supporting stochastic EP~\cite{li2015stochastic}. Differently, for binary classification, variational inference is adopted to derive a simple ELBO expressed as a one-dimensional Gaussian integral which can be calculated through Gauss-Hermite quadrature~\cite{hensman2015scalable}.\footnote{Due to the fast Gauss-Hermite quadrature with high precision, this ELBO is regarded as analytical.} This model has been further extended to multi-class classification~\cite{hensman2015mcmc}. Particularly, when using the logit likelihood, the P\`olya-Gamma data augmentation~\cite{polson2013bayesian} can be used such that the ELBO and the posterior are analytical in the augmented probability space~\cite{wenzel2018efficient}. This augmentation strategy has been recently extended to multi-class classification using logistic-softmax likelihood~\cite{galy2019multi}.\footnote{Different from the original softmax likelihood, this hybrid likelihood allows deriving analytical ELBO via a complex three-level augmentation.} Besides, a decoupled approach~\cite{cheng2017variational,salimbeni2018orthogonally} from the weight-space view removes the coupling between the mean and the covariance of a GP, resulting in lower complexity and an expressive prediction mean.

Though showing high scalability for handling big data, existing scalable GPCs derive analytical ELBO (i) using additional assumptions which may limit the representational capability~\cite{hernandez2016scalable, villacampa2017scalable}, and (ii) only for specific likelihoods, e.g., the step or probit likelihood~\cite{hensman2015scalable, villacampa2017scalable, wenzel2018efficient, galy2019multi, salimbeni2018orthogonally}.

Hence, this article proposes a unifying scalable GPC framework which accommodates various likelihoods without additional assumptions. Specifically, by interpreting the GPC as a noisy model analogous to GPR, we describe the step and (multinomial) probit/logit likelihoods over a general Gaussian error, and the softmax likelihood over a Gumbel error. Thereafter, we augment the probability space for (i) the GPCs using \textit{step} and (multinomial) \textit{probit/logit} likelihoods via the internal variables; and \textit{particularly}, (ii) the GPC using \textit{softmax} likelihood via the noises themselves. This leads to scalable GPCs with analytical ELBO by using variational inference. We empirically demonstrate the superiority of our GPCs on extensive binary/multi-class classification tasks with up to two million data points. Python implementations built upon the GPflow library~\cite{matthews2017gpflow} are available at~\url{https://github.com/LiuHaiTao01/GPCnoise}.

The reminder of this article is organized as follows. We introduce the proposed scalable binary/multi-class GPCs with additive noise in sections~\ref{sec_binary_GPC} and~\ref{sec_multiclass_GPC}, respectively. Thereafter, section~\ref{sec_results} conducts numerical experiments to assess the performance of proposed GPCs against state-of-the-art GPCs. Finally, section~\ref{sec_conclusion} offers concluding remarks.

\section{Binary GPCs with additive noise}
\label{sec_binary_GPC}
\subsection{Interpreting binary GPCs with additive noise}
\begin{figure}[t!]
	\centering
	\includegraphics[width=0.45\textwidth]{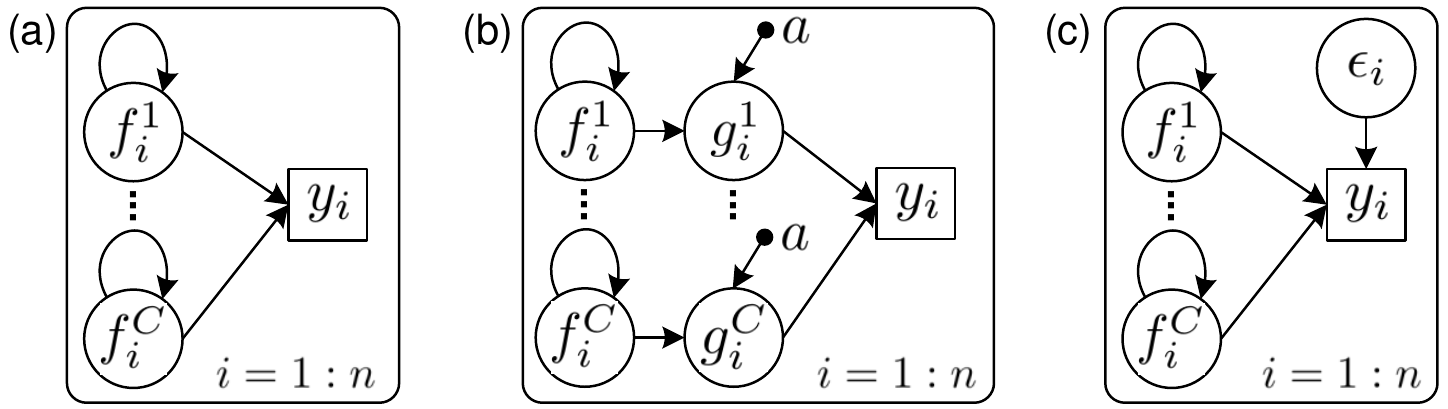}
	\caption{(a) Illustration of the GPC where for binary case $C=1$ and for multi-class case $C > 2$. (b) The GPC with additive Gaussian noise $\epsilon_i^c \in \mathcal{N}(0,a)$. This model is augmented by introducing the internal function $g$, which offers a unifying description and helps derive closed-form ELBO. By marginalizing $g$ out and varying over $a$, we recover the GPC in (a) using step and (multinomial) probit/logit likelihoods. (c) Particularly, for multi-class GPC using softmax likelihood, instead of introducing $g$, we directly use the Gumbel error $\epsilon_i$ to augment the probability space for deriving analytical ELBO.}
	\label{fig_GPCnoise}
\end{figure}

For the binary classification with $y \in \{-1,1\}$, the GPC model in Fig.~\ref{fig_GPCnoise}(a) is usually expressed as
\begin{align}
\label{eq_binary_GPC}
f(\mathbf{x}) \sim \mathcal{GP}(0, k(\mathbf{x}, \mathbf{x}')), \quad p(y|f) = \pi(yf),
\end{align}
where the likelihood $p(y|f)$ employs an inverse link function $\pi(.) \in [0,1]$ to squash the latent function $f$ into the class probability space. Commonly used $\pi(.)$ for binary GPC includes
\begin{equation}
\label{eq_pi_binary}
\begin{aligned}
\mathrm{step:} \ \pi(z) &= H(z), \\ \mathrm{probit:} \ \pi(z) &= \int_{-\infty}^z \mathcal{N}(\tau|0,1) d\tau, \\
\mathrm{logit:} \ \pi(z) &= (1 + \exp(-z))^{-1},
\end{aligned}
\end{equation}
where $H(z)=1$ when $z>0$; otherwise, $H(z)=0$.

It is known that similar to the GPR $y(\mathbf{x}) = f(\mathbf{x}) + \epsilon(\mathbf{x})$, the GPC in~\eqref{eq_binary_GPC} can also be interpreted as a noisy model. Specifically, we introduce a GPR-like internal latent function $g$ and the internal step likelihood, resulting in
\begin{equation}
\label{eq_binary_GPC_noise}
g(\mathbf{x}) = f(\mathbf{x}) + \epsilon(\mathbf{x}), \quad p(y|g) = H(yg),
\end{equation}
where $\epsilon(\mathbf{x})$ is an independent and identically distributed (\textit{i.i.d.}) noise which follows a distribution with probability density function (PDF) $\phi(.)$ and cumulative distribution function (CDF) $\Phi(.)$. The additional variable $g$ in~\eqref{eq_binary_GPC_noise} augments the probability space, and incurs the conditional independence $y \perp f | g$, which is crucial for deriving the analytical ELBO below. 

The conventional likelihoods in~\eqref{eq_pi_binary} can be recovered by marginalizing the internal $g$ out as\footnote{Eq.~\eqref{eq_lik_noise} holds only when $\epsilon$ follows a symmetric distribution, i.e., $\Phi(x) = 1-\Phi(-x)$. Otherwise, we have $p(y=1|f)=1-\Phi(-f)$ and $p(y=-1|f)=\Phi(-f)$.}
\begin{align}
\label{eq_lik_noise}
p(y|f) = \int p(y|g) p(g|f) dg = \Phi(yf).
\end{align}
It is observed that (i) when $\epsilon$ follows the Dirac-delta distribution $\phi_{\mathcal{D}}(\epsilon)$ which is zero everywhere except at the origin where it is infinite, Eq.~\eqref{eq_lik_noise} recovers the step likelihood; (ii) when $\epsilon$ follows the normal distribution $\phi_{\mathcal{N}}(\epsilon| 0,1)$, Eq.~\eqref{eq_lik_noise} recovers the probit likelihood; and finally, (iii) when $\epsilon$ follows the standard logistic distribution $\phi_{\mathcal{L}}(\epsilon| 0,1)$, Eq.~\eqref{eq_lik_noise} recovers the logit likelihood.

More interestingly, we could use a general Gaussian error $\epsilon \sim \mathcal{N}(0, a)$ to describe the above three kinds of errors in a unifying way, since all of them are \textit{symmetric}, \textit{bell-shaped} distributions. It is observed that (i) when $a \rightarrow 0$, the Gaussian error degenerates to the Dirac-delta error;\footnote{Directly using $a=0$ will incur numerical issue for $\phi_{\mathcal{N}}(.)$ and $\Phi_{\mathcal{N}}(.)$. But this issue can be sidestepped in the GPC model presented below.} (ii) when $a = 1$, the Gaussian error is equivalent to the normal error; and (iii) when $a = 2.897$, the Gaussian error produces a well CDF approximation to that of the logistic error with the maximum difference as 0.009, see Fig.~\ref{fig_errors} in Appendix~\ref{app_gaussian_approx}. Bowling and Khasawneh~\cite{bowling2009logistic} proposed using the logistic function to approximate the normal CDF $\Phi_{\mathcal{N}}$. Inversely, we here use the Gaussian error to approximate the logistic error. Note that the optimal $a$ for logistic error is derived in terms of CDF rather than PDF since we focus on the approximation quality of the likelihood~\eqref{eq_lik_noise}.

Now the internal model $g(\mathbf{x}) = f(\mathbf{x}) + \mathcal{N}(\epsilon|0, a)$ with additive Gaussian noise, as depicted in Fig.~\ref{fig_GPCnoise}(b), helps (i) describe the binary GPCs using different likelihoods uniformly; and moreover, (ii) derive unifying and analytical ELBO and prediction in section~\ref{sec_unifying_binary_GPC}.

\subsection{Binary GPCs using step, probit and logit likelihoods}
\label{sec_unifying_binary_GPC}
\subsubsection{Evidence lower bound}
Different from~\eqref{eq_binary_GPC_noise}, we here introduce a robust parameter $\delta$ like~\cite{de2016scalable}, resulting in the likelihoods
\begin{equation}
\label{eq_robust_step}
\begin{aligned}
p(y|g) &= (1-2\delta)H(yg) + \delta, \\ 
p(y|f) &= (1-2\delta)\Phi_{\mathcal{N}}\left(\frac{yf}{\sqrt{a}}\right) + \delta.
\end{aligned}
\end{equation}
The parameter $\delta$, which is pretty small (e.g., $10^{-3}$), (i) prevents the ELBO below from being infinite, which is caused by the logarithm form $\log p(y|g)$; and (ii) gives a degree of robustness to outliers. Note that with $\delta$, $p(y|g)$ is still a distribution since we have $p(y|g) = \{1-\delta, \delta\}$.

Given the data $\{\mathbf{X}, \mathbf{y}\}$ we have the joint prior $p(\mathbf{f}) = \mathcal{N}(\mathbf{f}|\mathbf{0}, \mathbf{K}_{nn})$ where $\mathbf{f} = (f_1, \cdots, f_n)^{\mathsf{T}}$ and $[\mathbf{K}_{nn}]_{ij} = k(\mathbf{x}_i, \mathbf{x}_j)$. Thereafter, we have the likelihoods $p(\mathbf{g}|\mathbf{f}) = \mathcal{N}(\mathbf{g}|\mathbf{f}, a\mathbf{I})$ and $p(\mathbf{y}|\mathbf{g}) = \prod_{i=1}^n p(y_i|g_i)$. To improve the scalability for dealing with big data, we consider $m$ inducing variables $\mathbf{u}$, which follow the same GP prior $p(\mathbf{u}) = \mathcal{N}(\mathbf{u}|\mathbf{0}, \mathbf{K}_{mm})$, for the latent variables $\mathbf{f}$.\footnote{$\mathbf{u}$ is assumed to be a sufficient statistic for $\mathbf{f}$, i.e., for any $\mathbf{z}$ we have $p(\mathbf{z}|\mathbf{u}, \mathbf{f}) = p(\mathbf{z}|\mathbf{u})$.} In the statistical model, a central task is deriving the posterior $p(\mathbf{g}, \mathbf{f}, \mathbf{u} | \mathbf{y}) \propto p(\mathbf{y}|\mathbf{g}) p(\mathbf{g}|\mathbf{f}) p(\mathbf{f}|\mathbf{u})$ given the observations. It however is intractable for GPC due to the non-Gaussian likelihood $p(\mathbf{y}|\mathbf{g})$.

Instead, we introduce a variational distribution $q(\mathbf{g}, \mathbf{f}, \mathbf{u}) = p(\mathbf{g}|\mathbf{f}) p(\mathbf{f}|\mathbf{u}) q(\mathbf{u})$\footnote{According to~\cite{titsias2009variational}, we have $p(\mathbf{z}|\mathbf{u},\mathbf{y}) = p(\mathbf{z}|\mathbf{u})$. Hence, $q(\mathbf{g}, \mathbf{f}, \mathbf{u}) = p(\mathbf{g},\mathbf{f}| \mathbf{u}, \mathbf{y}) q(\mathbf{u}) = p(\mathbf{g},\mathbf{f}| \mathbf{u}) q(\mathbf{u}) = p(\mathbf{g}|\mathbf{f}) p(\mathbf{f}|\mathbf{u}) q(\mathbf{u})$.} to approximate the exact posterior $p(\mathbf{g}, \mathbf{f}, \mathbf{u} | \mathbf{y})$. We then minimize their KL divergence $\mathrm{KL}(q(\mathbf{g}, \mathbf{f}, \mathbf{u}) || p(\mathbf{g}, \mathbf{f}, \mathbf{u} | \mathbf{y})) = \log p(\mathbf{y}) - \mathcal{L}$, which is equivalent to maximizing the evidence lower bound (ELBO) $\mathcal{L}$ expressed as
\begin{equation}
\label{eq_elbo_binary_early}
\begin{aligned}
\mathcal{L} &= \left\langle \log \frac{p(\mathbf{y}, \mathbf{g}, \mathbf{f}, \mathbf{u})}{q(\mathbf{g}, \mathbf{f}, \mathbf{u})} \right\rangle_{q(\mathbf{g}, \mathbf{f}, \mathbf{u})} \\
&= \left\langle \log p(\mathbf{y}| \mathbf{g}) \right\rangle_{q(\mathbf{g})} - \mathrm{KL}(q(\mathbf{u}) || p(\mathbf{u})),
\end{aligned}
\end{equation}
where $\left\langle .\right\rangle_{q(.)}$ is the expectation over distribution $q(.)$, and $q(\mathbf{g}) = \int p(\mathbf{g}|\mathbf{f}) p(\mathbf{f}|\mathbf{u}) q(\mathbf{u}) d\mathbf{f}d\mathbf{u} = \int p(\mathbf{g}|\mathbf{f}) q(\mathbf{f}) d\mathbf{f}$.
For the GPR-like internal model $g(\mathbf{x}) = f(\mathbf{x}) + \mathcal{N}(\epsilon|0, a)$, we could derive an analytical posterior $q(\mathbf{g})$. Specifically, given that $q(\mathbf{u}) = \mathcal{N}(\mathbf{u}|\mathbf{m}, \mathbf{S})$, we have
\begin{equation}
\label{eq_qf_qg_binary}
\begin{aligned}
q(\mathbf{f}) &= \int p(\mathbf{f}|\mathbf{u}) q(\mathbf{u}) d\mathbf{u} = \mathcal{N}(\mathbf{f}|\bm{\mu}_f, \mathbf{\Sigma}_f), 
\\ q(\mathbf{g}) &= \int p(\mathbf{g}|\mathbf{f}) q(\mathbf{f}) d\mathbf{f} = \mathcal{N}(\mathbf{g}|\bm{\mu}_g, \mathbf{\Sigma}_g),
\end{aligned}
\end{equation}
where $\bm{\mu}_f = \mathbf{K}_{nm}\mathbf{K}_{mm}^{-1}\mathbf{m}$, $\mathbf{\Sigma}_f = \mathbf{K}_{nn} +\mathbf{K}_{nm}\mathbf{K}_{mm}^{-1}[\mathbf{S}\mathbf{K}_{mm}^{-1}-\mathbf{I}]\mathbf{K}_{mn}$, $\bm{\mu}_g = \bm{\mu}_f$ and $\mathbf{\Sigma}_g = a\mathbf{I} + \mathbf{\Sigma}_f$.
Inserting~\eqref{eq_qf_qg_binary} back into~\eqref{eq_elbo_binary_early}, we obtain a closed-form ELBO factorized over data points as
\begin{equation}
\label{eq_elbo_binary}
\begin{aligned}
\mathcal{L} =& \sum_{i=1}^{n} \left[ \log \left(\frac{1-\delta}{\delta} \right) \Phi_{\mathcal{N}} \left( \frac{y_i \mu_{f_i}}{\sqrt{a+\nu_{f_i}}} \right) + \log\delta \right] \\
&- \mathrm{KL}(q(\mathbf{u}) || p(\mathbf{u})),
\end{aligned}
\end{equation}
where $\mu_{f_i}=[\bm{\mu}_f]_i$ and $\nu_{f_i}=[\bm{\Sigma}_f]_{ii}$. The maximization of $\mathcal{L}$ permits inferring the variational parameters and hyperparameters simultaneously. Besides, the sum term in the right-hand side of $\mathcal{L}$ allows using efficient stochastic optimizer, e.g., the Adam~\cite{kingma2014adam}, for model training.\footnote{Particularly, the natural gradient descent (NGD) could be employed for optimizing the variational parameters $\mathbf{m}$ and $\mathbf{S}$. But in comparison to GPR, the NGD+Adam optimizer brings little benefits for classification~\cite{salimbeni2018natural,salimbeni2018orthogonally}.}

It is observed from~\eqref{eq_elbo_binary} that (i) when $a = 0$, we obtain the bound $\mathcal{L}$ for the binary GPC using step likelihood; (i) when $a = 1$, we obtain the binary GPC using probit likelihood; and (i) when $a = 2.897$, we obtain the binary GPC using logit likelihood. The unifying bound~\eqref{eq_elbo_binary} reveals that the binary GPCs using step, probit and logit likelihoods have similar behavior. 

\subsubsection{Prediction}
The prediction for $f_*$ at $\mathbf{x}_*$ is $p(f_*|\mathbf{y}) = \int p(f_*|\mathbf{f}) q(\mathbf{f}) d\mathbf{f} = \mathcal{N}(f_*|\mu_{f_*}, \nu_{f_*})$,
where $\mu_{f_*} = \mathbf{k}_{*m} \mathbf{K}_{mm}^{-1} \mathbf{m}$, and $\nu_{f_*} = k_{**} + \mathbf{k}_{*m}\mathbf{K}_{mm}^{-1}[\mathbf{S}\mathbf{K}_{mm}^{-1}-\mathbf{I}]\mathbf{k}_{m*}$. Similarly, we obtain the prediction for $g_*$ at $\mathbf{x}_*$ as $p(g_*|\mathbf{y}) = \int p(g_*|\mathbf{g}) q(\mathbf{g}) d\mathbf{g} = \mathcal{N}(g_*|\mu_{g_*}, \nu_{g_*})$, where $\mu_{g_*} = \mu_{f_*}$ and $\nu_{g_*} = \nu_{f_*} + a$. Finally, we have the closed-form class probability for $y_*=1$ as
\begin{equation}
\begin{aligned}
p(y_*=1|\mathbf{y}) &= \int p(y_*=1|g_*) p(g_*|\mathbf{y}) df_* \\
&= (1-2\delta) \Phi_{\mathcal{N}}\left( \frac{\mu_{f_*}}{\sqrt{a+\nu_{f_*}}} \right) + \delta.
\end{aligned}
\end{equation}

\subsection{Discussions} 
We would like to emphasize that the binary GPCs proposed in section~\ref{sec_unifying_binary_GPC} are different from that in section~3.2 of~\cite{hensman2015scalable}. Hensman et al.~\cite{hensman2015scalable} borrowed the bound $\mathcal{L}_g$ for $\log p(\mathbf{g})$ in~\cite{hensman2013gaussian}, which is equivalent to maximizing $\mathrm{KL}(q(\mathbf{f},\mathbf{u}|\mathbf{g})||p(\mathbf{f},\mathbf{u}|\mathbf{g}))$, and substituted it into the augmented joint distribution $p(\mathbf{y}, \mathbf{g}, \mathbf{f})$. This results in $\log p(\mathbf{y}) \ge \log \int p(\mathbf{y}| \mathbf{g}) \exp(\mathcal{L}_g) d\mathbf{g} = \mathcal{L}_{\mathrm{HS}}$. When using robust $p(\mathbf{y}|\mathbf{g})$, we have
\begin{equation}
\begin{aligned}
\mathcal{L}_{\mathrm{HS}} =& \sum_{i=1}^n \log((1-2\delta) \Phi_{\mathcal{N}}(y_i \mu_{f_i}) + \delta) - \frac{1}{2} \mathrm{tr}(\mathbf{\Sigma}_f) \\
&- \mathrm{KL}(q(\mathbf{u})||p(\mathbf{u})),
\end{aligned}
\end{equation}
which is different from our bound~\eqref{eq_elbo_binary}. Let $\mathbf{X}_m = \mathbf{X}$ so that $\mathbf{u} = \mathbf{f}$, $\mathcal{L}_{\mathrm{HS}}$ has a unique optimum where $\mathbf{m}_* = \mathbf{K}_{nn} \partial \alpha(\mathbf{m}) / \partial \mathbf{m}$ and $\mathbf{S}_* = (\mathbf{K}_{nn}^{-1} + \mathbf{I})^{-1}$ with $\alpha(\mathbf{m})$ being the first term in the right-hand side of $\mathcal{L}_{\mathrm{HS}}$. The ineffective estimations $\mathbf{m}_*$ and $\mathbf{S}_*$  are not due to the decoupling $y \bot f|g$. They occur since $\mathcal{L}_{\mathrm{HS}}$ \textit{seeks to infer} $q(\mathbf{f}|\mathbf{g})$ \textit{rather than the interested} $q(\mathbf{f}|\mathbf{y})$.
	
In contrast, our binary GPCs use VI to \textit{directly} approximate $\log p(\mathbf{y})$ via maximizing $\mathrm{KL}(q(\mathbf{g}, \mathbf{f},\mathbf{u}|\mathbf{y})||p(\mathbf{g}, \mathbf{f},\mathbf{u}|\mathbf{y})$. The improvement occurs that the internal variables $\mathbf{g}$ and the general Gaussian error help derive \textit{completely} analytical ELBOs for binary GPCs using step, probit and logit likelihoods. Let $\mathbf{X}_m = \mathbf{X}$, our bound $\mathcal{L}$ in~\eqref{eq_elbo_binary} has a unique optimum where $\mathbf{m}_* = \mathbf{K}_{nn} \partial \beta(a, \mathbf{m}, \mathbf{S}) / \partial \mathbf{m}$ and $\mathbf{S}_* = (\mathbf{K}_{nn}^{-1} - 2 \times \partial \beta(a, \mathbf{m}, \mathbf{S}) / \partial \mathbf{S})^{-1}$ with $\beta$ being the first term in the right-hand side of $\mathcal{L}$. This optimum, which is similar to that of the KL method rather than the Laplace approximation in~\cite{nickisch2008approximations}, is more informative than that of $\mathcal{L}_{\mathrm{HS}}$.

\section{Multi-class GPCs with additive noise}
\label{sec_multiclass_GPC}
\subsection{Interpreting multi-class GPCs with additive noise}
The more complicated multi-class GPC with the label $y \in \{1, \cdots, C\}$, $C > 2$, is expressed by introducing the internals $\{g^c\}_{c=1}^C$ for each of the $C$ classes as
\begin{equation}
\label{eq_multiclass_GPC_noise}
\begin{aligned}
f^c(\mathbf{x}) &\sim \mathcal{GP}(0, k^c(\mathbf{x}, \mathbf{x}')), \\
g^c(\mathbf{x}) &= f^c(\mathbf{x}) + \epsilon^c(\mathbf{x}), \\ 
y(\mathbf{x}) &= \underset{c}{\mathrm{argmax}} g^c(\mathbf{x}),
\end{aligned}
\end{equation}
where $f^c$ and $g^c$ are independent latent functions, and $\epsilon^c$ is the \textit{i.i.d.} noise for class $c$ ($1 \le c \le C$). For notations, we first define $\mathbf{f}_i = (f^1(\mathbf{x}_i), \cdots, f^C(\mathbf{x}_i))^{\mathsf{T}} \in R^C$, $\mathbf{f}^c = (f^c(\mathbf{x}_1), \cdots, f^c(\mathbf{x}_n))^{\mathsf{T}} \in R^n$, $\mathbf{f} = (\mathbf{f}^1, \cdots, \mathbf{f}^C) \in R^{n \times C}$, $\mathbf{g}_i = (g^1(\mathbf{x}_i), \cdots, g^C(\mathbf{x}_i))^{\mathsf{T}} \in R^C$, $\mathbf{g}^c = (g^c(\mathbf{x}_1), \cdots, g^c(\mathbf{x}_n))^{\mathsf{T}} \in R^n$, and $\mathbf{g} = (\mathbf{g}^1, \cdots, \mathbf{g}^C) \in R^{n \times C}$. 
We adopt the internal step likelihood
\begin{align}
p(y_i|\mathbf{g}_i) = \prod_{c \ne y_i} H(g_i^{y_i} - g_i^c ) = \prod_{c \ne y_i} H(f_i^{y_i} + \epsilon_i^{y_i} - f_i^c - \epsilon_i^c).
\end{align}
Thereafter, by integrating $\mathbf{g}_i$ out, we obtain the conventional likelihood\footnote{We omit the superscript in $\epsilon^{y_i}_i$ since Eq.~\eqref{eq_lik_multiclass} has removed the dependency on other error terms $\{\epsilon^{c}_i\}_{c \ne y_i}$.}
\begin{equation}
\label{eq_lik_multiclass}
\begin{aligned}
p(y_i|\mathbf{f}_i) &= \int p(y_i|\mathbf{g}_i) p(\mathbf{g}_i|\mathbf{f}_i) d\mathbf{g}_i \\
&=\int \phi(\epsilon_i) \prod_{c \ne y_i} \Phi(\epsilon_i+f_i^{y_i}-f_i^c) d\epsilon_i.
\end{aligned}
\end{equation}

It is observed that (i) when $\epsilon_i \sim \phi_{\mathcal{D}}(\epsilon)$, Eq.~\eqref{eq_lik_multiclass} recovers the multi-class step likelihood; (ii) when $\epsilon_i \sim \phi_{\mathcal{N}}(\epsilon|0,1)$, Eq.~\eqref{eq_lik_multiclass} recovers the multinomial probit likelihood; (iii) when $\epsilon_i \sim \phi_{\mathcal{L}}(\epsilon|0,1)$, Eq.~\eqref{eq_lik_multiclass} recovers the multinomial logit likelihood; and finally, (iv) when using the Gumbel error $\epsilon_i \sim \phi_{\mathcal{G}}(\epsilon|0,1)$, Eq.~\eqref{eq_lik_multiclass} recovers the softmax likelihood, i.e.,
\begin{align}
\label{eq_sm}
p(y_i|\mathbf{f}_i) = \frac{\exp(f_i^{y_i})}{ \sum_{c=1}^C \exp(f^c_i)}.
\end{align}

Similarly, as depicted in Fig.~\ref{fig_GPCnoise}(b), we can employ a Gaussian error $\epsilon \sim \mathcal{N}(0, a)$ to describe the first three symmetric errors with $a \rightarrow 0$, $a=1$, and $a=2.897$, respectively. But this is not the case for the \textit{asymmetric} Gumbel error, see Fig.~\ref{fig_errors} in Appendix~\ref{app_gaussian_approx}. Hence, section~\ref{sec_unifying_multiclass_GPC} introduces the sparse multi-class GPCs using step and multinomial probit/logit likelihoods in a unifying framework; \textit{particularly}, section~\ref{sec_GPCsm} introduces the sparse multi-class GPC using softmax likelihood.

\subsection{Multi-class GPCs using step and multinomial probit/logit likelihoods}
\label{sec_unifying_multiclass_GPC}
\subsubsection{Evidence lower bound}
Similar to~\eqref{eq_robust_step}, we employ the robust likelihoods as
\begin{equation}
\label{eq_multiclass_robust_step}
\begin{aligned}
p(y_i|\mathbf{g}_i) &= \left(1-\frac{C}{C-1}\delta \right) \prod_{c\ne y_i} H(g_i^{y_i}-g_i^c) + \frac{\delta}{C-1}, \\ 
p(y_i|\mathbf{f}_i) &=  (1-\frac{C}{C-1}\delta)S_0 + \frac{\delta}{C-1},
\end{aligned}
\end{equation}
where, given that $p(\mathbf{g}_i|\mathbf{f}_i) = \mathcal{N}(\mathbf{g}_i| \mathbf{f}_i, a\mathbf{I})$, $
S_0 = \mathbb{E}_{\epsilon_i \sim \mathcal{N}(0,a)} \left[\prod_{c \ne y_i} \Phi_{\mathcal{N}} \left( \frac{\epsilon_i + f^{y_i}_i - f^c_i}{\sqrt{a}} \right) \right]$.

Again, by introducing the independent inducing set $\mathbf{u}^c$ for $\mathbf{f}^c$, $1 \le c \le C$, the ELBO writes
\begin{equation}
\label{eq_elbo_multiclass_original}
\begin{aligned}
\mathcal{L} &= \left\langle   \log \frac{p(\mathbf{y}, \mathbf{g}, \mathbf{f}, \mathbf{u})}{q(\mathbf{g}, \mathbf{f}, \mathbf{u})} \right\rangle_{q(\mathbf{g}, \mathbf{f}, \mathbf{u})} \\
&= \left\langle \log p(\mathbf{y}| \mathbf{g}) \right\rangle_{q(\mathbf{g})} - \sum_{c=1}^C\mathrm{KL}(q(\mathbf{u}^c) || p(\mathbf{u}^c)).
\end{aligned}
\end{equation}
Let $q(\mathbf{u}^c) = \mathcal{N}(\mathbf{u}^c| \mathbf{m}^c, \mathbf{S}^c)$, we have
\begin{equation}
\label{eq_qf_qg_multiclass}
\begin{aligned}
q(\mathbf{f}) &= \int p(\mathbf{f}|\mathbf{u}) q(\mathbf{u}) d\mathbf{u} = \prod_{c=1}^C \mathcal{N}(\mathbf{f}^c|\bm{\mu}^c_f, \mathbf{\Sigma}^c_f),\\ 
q(\mathbf{g}) &= \int p(\mathbf{g}|\mathbf{f}) q(\mathbf{f}) d\mathbf{f} = \prod_{c=1}^C \mathcal{N}(\mathbf{g}^c|\bm{\mu}^c_g, \mathbf{\Sigma}^c_g),
\end{aligned}
\end{equation}
where $\bm{\mu}^c_f = \mathbf{K}^c_{nm}(\mathbf{K}_{mm}^c)^{-1}\mathbf{m}^c$, $\mathbf{\Sigma}_f^c = \mathbf{K}^c_{nn} +\mathbf{K}^c_{nm}(\mathbf{K}_{mm}^c)^{-1} [\mathbf{S}^c(\mathbf{K}_{mm}^c)^{-1} - \mathbf{I}] \mathbf{K}^c_{mn}$, $\bm{\mu}^c_g = \bm{\mu}^c_f$ and $\mathbf{\Sigma}^c_g = a\mathbf{I} + \mathbf{\Sigma}^c_f$. Inserting~\eqref{eq_qf_qg_multiclass} back into~\eqref{eq_elbo_multiclass_original}, we have a factorized ELBO as
\begin{equation}
\label{eq_elbo_multiclass}
\begin{aligned}
\mathcal{L} =& \sum_{i=1}^n \left[\log(1-\delta)S_i + \log\left(\frac{\delta}{C-1}\right)(1-S_i)\right] \\
&- \sum_{c=1}^C \mathrm{KL}(q(\mathbf{u}^c) || p(\mathbf{u}^c)).
\end{aligned}
\end{equation}
where $
S_i = \mathbb{E}_{g_i^{y_i} \sim \mathcal{N}(\mu_{f_i}^{y_i}, a+\nu_{f_i}^{y_i})} \left[\prod_{c \ne y_i} \Phi_{\mathcal{N}} \left(\frac{g_i^{y_i}-\mu_{f_i}^c}{\sqrt{a + \nu_{f_i}^c}} \right) \right]$.
The term $S_i$ is interpreted as the probability that the function value corresponding to the
observed class $y_i$ is larger than the others at $\mathbf{x}_i$. Note that this \textit{one dimensional Gaussian integral} can be evaluated using fast Gauss-Hermite quadrature with high precision.

The unifying ELBO~\eqref{eq_elbo_multiclass} herein describes the multi-class GPCs using step likelihood ($a=0$), multinomial probit likelihood ($a=1$) and multinomial logit likelihood ($a=2.897$). Note that when $a=0$, the model is equivalent to the one presented in~\cite{de2016scalable}.

\subsubsection{Prediction}
Finally, the prediction for $\mathbf{f}_*=(f_*^1,\cdots,f_*^C)^{\mathsf{T}}$ at $\mathbf{x}_*$ is $p(\mathbf{f}_*|\mathbf{y}) = \int p(\mathbf{f}_*|\mathbf{f}) q(\mathbf{f}) d\mathbf{f} = \prod_{c=1}^C \mathcal{N}(f_*^c|\mu_{f_*}^c, \nu_{f_*}^c)$,
where $\mu_{f_*}^c = \mathbf{k}^c_{*m} (\mathbf{K}_{mm}^c)^{-1} \mathbf{m}^c$ and $\nu^c_{f_*} = k^c_{**} + \mathbf{k}^c_{*m} (\mathbf{K}_{mm}^c)^{-1}[\mathbf{S}^c(\mathbf{K}_{mm}^c)^{-1}-\mathbf{I}]\mathbf{k}_{m*}^c$.
Similarly, we have the prediction for $\mathbf{g}_*$ at $\mathbf{x}_*$ as $p(\mathbf{g}_*|\mathbf{y}) = \int p(\mathbf{g}_*|\mathbf{g}) q(\mathbf{g}) d\mathbf{g} = \prod_{c=1}^C \mathcal{N}(g_*^c|\mu_{g_*}^c, \nu_{g_*}^c)$, where $\mu_{g_*}^c = \mu_{f_*}^c$ and $\nu_{g_*}^c = \nu_{f_*}^c + a$.
Finally, we have
\begin{equation}
\begin{aligned}
p(y_*|\mathbf{y}) &= \int p(y_*|\mathbf{g}_*) p(\mathbf{g}_*|\mathbf{y}) d\mathbf{f}_* \\
&= (1-\delta)S_* + \frac{\delta}{C-1}(1-S_*),
\end{aligned}
\end{equation}
where $S_* = \mathbb{E}_{g_*^{y_*} \sim \mathcal{N}(\mu_{f_*}^{y_*}, a+\nu_{f_*}^{y_*})} \left[\prod_{c \ne y_*} \Phi_{\mathcal{N}} \left(\frac{g_*^{y_*}-\mu_{f_*}^c}{\sqrt{a + \nu_{f_*}^c}} \right) \right]$.

\subsection{A possible generalization?}
It is found that the \textit{unifying} ELBOs in~\eqref{eq_elbo_binary} and~\eqref{eq_elbo_multiclass} respectively recover the binary and multi-class GPCs using step and (multinomial) probit/logit likelihoods by varying over the Gaussian noise variance $a$. It indicates that the GPCs using these three \textit{symmetric} likelihoods would produce similar results, which will be verified in the numerical experiments in section~\ref{sec_results}.

Besides, the parameter $a$ inspires us to think of a more general GPC model, like standard GPR, by treating the variance $a$ as a hyperparameter and inferring it from data. However, as discussed in Appendix~\ref{app_opt_a}, the ELBOs in~\eqref{eq_elbo_binary} and~\eqref{eq_elbo_multiclass} increase with decreasing $a$. That means, $\mathcal{L}$ arrives at the maximum with $a = 0$. This is because the internal likelihood $p(\mathbf{y}|\mathbf{g})$ does not take into account the noise variance. 

We name the GPCs using step and (multinomial) probit/logit likelihoods as GPC-I, GPC-II and GPC-III, respectively. It is found that given the same configuration of hyperparameters, the ELBO satisfies $\mathcal{L}_{\mathrm{I}} > \mathcal{L}_{\mathrm{II}} > \mathcal{L}_{\mathrm{III}}$, which again verifies that the noise variance $a$ is not a proper hyperparameter.

\subsection{Multi-class GPC using softmax likelihood}
\label{sec_GPCsm}
\subsubsection{Evidence lower bound}
According to~\eqref{eq_lik_multiclass}, the softmax likelihood used in the noisy multi-class GPC is equivalent to
\begin{align}
\label{eq_sm_lik_noise}
p(y_i|\mathbf{f}_i) =\int \phi_{\mathcal{G}}(\epsilon_i) \prod_{c \ne y_i} \Phi_{\mathcal{G}}(\epsilon_i+f_i^{y_i}-f_i^c) d\epsilon_i,
\end{align}
where for the standard Gumbel error, the PDF is $\phi_{\mathcal{G}}(x)=\exp(-x-e^{-x})$ and the CDF is $\Phi_{\mathcal{G}}(x)=\exp(-e^{-x})$. Instead of introducing the internal $\mathbf{g}$, we here directly use the noise variables $\bm{\epsilon} = \{\epsilon_i\}_{i=1}^n$ to augment the probability space due to the expectation form in~\eqref{eq_sm_lik_noise}, as depicted in Fig.~\ref{fig_GPCnoise}(c). 
Hence, the augmented model with $\epsilon_i$ has the following conditional distributions as
\begin{equation}
\begin{aligned}
p(y_i,\epsilon_i|\mathbf{f}_i) &= \phi_{\mathcal{G}}(\epsilon_i) \prod_{c \ne y_i} \Phi_{\mathcal{G}}(\epsilon_i+f_i^{y_i}-f_i^c), \\ 
p(y_i|\epsilon_i,\mathbf{f}_i) &= \prod_{c \ne y_i} \Phi_{\mathcal{G}}(\epsilon_i+f_i^{y_i}-f_i^c).
\end{aligned}
\end{equation}

By introducing the inducing set $\mathbf{u}^c$ for $\mathbf{f}^c$, $1 \le c \le C$, we derive the ELBO as
\begin{equation} \label{eq_elbo_mgpc_aug}
\begin{aligned}
\mathcal{L} =& \left\langle   \log \frac{p(\mathbf{y}, \mathbf{f}, \mathbf{u}, \bm{\epsilon})}{q(\mathbf{f}, \mathbf{u}, \bm{\epsilon})} \right\rangle_{q(\mathbf{f}, \mathbf{u}, \bm{\epsilon})} \\
=& \sum_i^n \left\langle \log p(y_i| \mathbf{f}_i, \epsilon_i) \right\rangle_{q(\mathbf{f}_i)q(\epsilon_i|\mathbf{f}_i)} - \sum_{i=1}^n\mathrm{KL}(q(\epsilon_i|\mathbf{f}_i) || p(\epsilon_i)) \\
&-  \sum_{c=1}^C\mathrm{KL}(q(\mathbf{u}^c|\mathbf{y}) || p(\mathbf{u}^c)),
\end{aligned}
\end{equation}
where $q(\mathbf{f}, \mathbf{u}, \bm{\epsilon}) = p(\mathbf{f}|\mathbf{u}) q(\mathbf{u}) q(\bm{\epsilon}|\mathbf{f})$; $q(\mathbf{f}_i) = \prod_{c=1}^C q(f_i^c) = \prod_{c=1}^C \mathcal{N}(f_i^c| \mu_{f_i}^c, \nu_{f_i}^c)$ with $\mu_{f_i}^c=[\bm{\mu}^c_f]_i$ and $\nu_{f_i}^c=[\mathbf{\Sigma}^c_f]_{ii}$; and $q(\epsilon_i|\mathbf{f}_i)$ approximates the posterior $p(\epsilon_i|y_i,\mathbf{f}_i)$, which follows the exact expression as
\begin{equation}
\begin{aligned}
p(\epsilon_i|y_i,\mathbf{f}_i) &\propto p(y_i|\mathbf{f}_i,\epsilon_i)p(\epsilon_i) \\
&=\phi_{\mathcal{G}}(\epsilon_i) \prod_{c \ne y_i} \Phi_{\mathcal{G}}(\epsilon_i+f^{y_i}_i-f_i^c) \\
&= \exp\left(-\epsilon_i-\left(1+\sum_{c\ne y_i}e^{f_i^c-f_i^{y_i}}\right)e^{-\epsilon_i} \right) \\
&\overset{\mathsf{c}}{=} \mathrm{Gumbel}\left(\epsilon_i|\log\theta^*_i,1\right),
\end{aligned}
\end{equation}
where $\theta^*_i=1+\sum_{c\ne y_i}e^{f_i^c-f_i^{y_i}}=\sum_{c=1}^C e^{f_i^c-f_i^{y_i}}$. Note that though the optimal $p(\epsilon_i|y_i,\mathbf{f}_i)$ has an analytic form, directly using $p(\epsilon_i|y_i,\mathbf{f}_i)$ will make the ELBO intractable. Hence, we adopt a general distribution $q(\epsilon_i|\mathbf{f}_i)=\mathrm{Gumbel}(\epsilon_i|\log\theta_i,1)$ which satisfies $\theta_i > 1$ and already contains the optimal distribution.

Thereafter, the closed-form ELBO, which is detailed in Appendix~\ref{app_elbo_softmax}, is reorganized as, 
\begin{align}
\mathcal{L} = \sum_{i=1}^n \left\{
-\frac{1}{\theta_i} \mathcal{P}_i - \log\theta_i-\frac{1}{\theta_i}+1 
\right\} -\sum_{c=1}^C\mathrm{KL}(q(\mathbf{u}^c|\mathbf{y}) || p(\mathbf{u}^c)),
\end{align}
where $\mathcal{P}_i = \exp \left(\frac{\nu_{f_i}^{y_i}}{2}-\mu_{f_i}^{y_i}\right) \sum_{c \ne y_i} \exp \left(\frac{\nu_{f_i}^c}{2}+\mu_{f_i}^c\right)$. Furthermore, in order to obtain a tight bound, let the derivative of $\mathcal{L}$ w.r.t. $\theta_i$
\begin{equation*}
\frac{\partial \mathcal{L}}{\partial \theta_i} = \frac{1}{\theta_i^2} (\mathcal{P}_i +1) - \frac{1}{\theta_i}
\end{equation*}
to be zero, we have the optimal value $\theta_i^* = \mathcal{P}_i +1$.
Substituting $\theta_i^*$ into $\mathcal{L}$, we have
\begin{equation}
\label{eq_ELBO_sm}
\mathcal{L} = -\sum_{i=1}^n \log (\mathcal{P}_i +1) -\sum_{c=1}^C\mathrm{KL}(q(\mathbf{u}^c|\mathbf{y}) || p(\mathbf{u}^c)).
\end{equation}
We here name the GPC using softmax likelihood as GPCsm.

\subsubsection{Prediction}
To predict the latent function values $\mathbf{f}_*$ at the test point $\mathbf{x}_*$, we substitute the approximate posteriors into the predictive distribution $p(\mathbf{f}_*|\mathbf{y}) = \int p(\mathbf{f}_*|\mathbf{u}) q(\mathbf{u}) d\mathbf{u} = \prod_{c=1}^C \mathcal{N}(f_*^c|\mu_{f_*}^c,\nu^c_{f_*})$. Thereafter, the distribution of the test label is computed as
\begin{equation}
\begin{aligned}
p(y_*|\mathbf{y}) &= \int p(y_*|\mathbf{f}_*) p(\mathbf{f}_*|\mathbf{y}) d\mathbf{f}_* \\
&= \int \frac{\exp(f_*^{y_*})}{\sum_{c=1}^C \exp(f_*^c)} \prod_{c=1}^C \mathcal{N}(f_*^c|\mu_{f_*}^c,\nu^c_{f_*}) d\mathbf{f}_*.
\end{aligned}
\end{equation}
The resulting integral however is intractable. Simply, we estimate the mean prediction $p(y_*|\mathbf{y})$ via markov chain monte carlo (MCMC) by drawing samples from the Gaussian $p(\mathbf{f}_*|\mathbf{y})$. Due to the independent latent functions, we draw samples from the posterior $p(\mathbf{f}_*|\mathbf{y})$ as follows: let $\mathbf{t} \sim \mathcal{N}(\mathbf{0}, \mathbf{I})$, we have the sample $\mathbf{x}_{\diamond} = \bm{\mu}_{f_*} + \sqrt{\bm{\nu}_{f_*}} \circ \mathbf{t}$ where the symbol $\circ$ represents the point-wise product.

\section{Numerical experiments}
\label{sec_results}
This section verifies the performance of the proposed GPCs (GPC-I, GPC-II, GPC-III and GPCsm) on multiple binary/multi-class classification tasks. We compare them against state-of-the-art scalable GPCs including (i) the binary/multi-class GPC using EP (GPCep)~\cite{hernandez2016scalable, villacampa2017scalable} (R codes available at \url{http://proceedings.mlr.press/v51/hernandez-lobato16.html} and \url{http://proceedings.mlr.press/v70/villacampa-calvo17a.html}), (ii) the binary/multi-class GPC using data augmentation (GPCaug)~\cite{wenzel2018efficient, galy2019multi} (Julia codes available at~\url{https://github.com/theogf/AugmentedGaussianProcesses.jl}), and (iii) the binary/multi-class GPC using orthogonally decoupled basis functions (ORTH)~\cite{salimbeni2018orthogonally} (Python codes available at~\url{https://github.com/hughsalimbeni/orth_decoupled_var_gps}). We run the experiments on a Linux workstation with eight 3.20 GHz cores, nvidia GTX1080Ti, and 32GB memory.

Table~\ref{tab_characteris_GPCs} summarizes the capabilities of existing scalable GPCs for handling various likelihoods. Specifically, the GPCep and ORTH employ the probit likelihood for binary case and the step likelihood for multi-class case; and the GPCaug uses the logit likelihood for binary case and the logistic-softmax likelihood\footnote{Similar to the softmax~\eqref{eq_sm}, this likelihood expresses as $p(y_i|\mathbf{f}_i) = \sigma(f_i^{y_i}) / \sum_{c=1}^C \sigma(f^c_i)$ with $\sigma(z) = (1 + \exp(-z))^{-1}$.} for multi-class case.

\begin{table}[h]
	\caption{The capabilities of existing scalable GPCs for handling various likelihoods. The symbol ``$\mathrm{b}$'' represents binary case and ``$\mathrm{m}$'' represents multi-class case. Note that the $\mathrm{GPCaug}$ employs a logistic-softmax likelihood for multi-class case.}
	\centering
	\begin{tabular}{lllll}%
		\hline
		&  Step lik. & Probit lik. & Logit lik. & Softmax lik. \\
		\hline             
		GPCep~\cite{hernandez2016scalable, villacampa2017scalable} & \cmark (m) & \cmark (b)  & \xmark  & \xmark \\
		GPCaug~\cite{wenzel2018efficient, galy2019multi} & \xmark & \xmark & \cmark (b) & \cmark (m) \\
		ORTH~\cite{salimbeni2018orthogonally} & \cmark (m) & \cmark (b) & \xmark  & \xmark \\
		Ours & \cmark (b, m) & \cmark (b, m) & \cmark (b, m) & \cmark (m) \\
		\hline
	\end{tabular}
	\label{tab_characteris_GPCs}
\end{table}

\subsection{Comparative results}
\subsubsection{Toy examples} We showcase the proposed GPCs on two illustrative binary and multi-class cases. The binary case is the \texttt{banana} dataset used in~\cite{hensman2015scalable}; the synthetic three-class case samples three latent functions from a GP prior, and applies the rule $y(\mathbf{x}) = \mathrm{argmax}_c f^c(\mathbf{x})$ to them. For the two classification datasets, we use $m = 32$ and the Mat\'{e}rn32 kernel for GPCs. The models are trained using the Adam optimizer~\cite{kingma2014adam} over 5000 iterations. 

\begin{figure}[t!]
	\centering
	\begin{subfigure}
		\centering
		\includegraphics[width=0.45\textwidth]{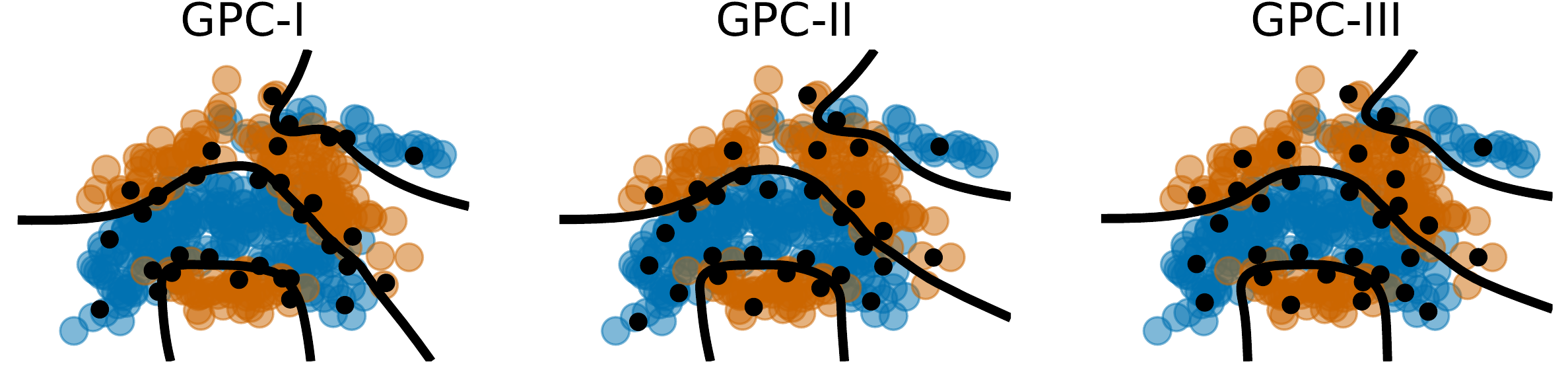}
	\end{subfigure}%
	\begin{subfigure}
		\centering
		\includegraphics[width=0.45\textwidth]{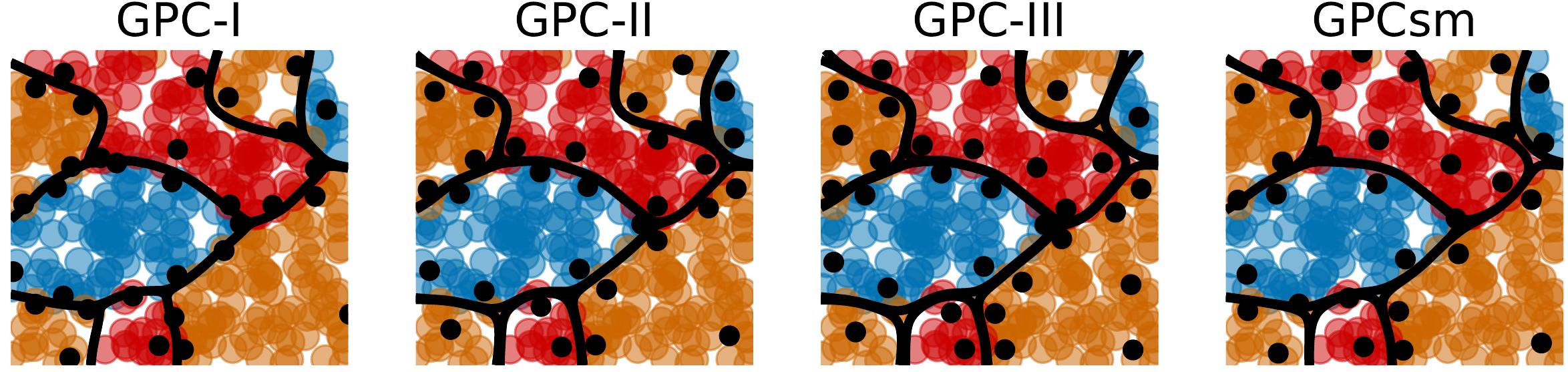}
	\end{subfigure} 
	\caption{Illustration of the proposed GPCs on a binary case (top row) and a three-class case (bottom row). The black curves represent the contours at which the predicted class probability is 0.5. The black circles represent the optimized locations of inducing points.}
	\label{Fig_toy} 
\end{figure}

As shown in Fig.~\ref{Fig_toy}, it is observed that (i) the proposed GPCs well classify the two datasets; and (ii) the variational inference framework pushes the optimized inducing points towards the decision boundaries, which is similar to~\cite{hensman2015scalable} but different from~\cite{hernandez2016scalable, villacampa2017scalable}.\footnote{The FITC assumption in~\cite{hernandez2016scalable, villacampa2017scalable} incurs overlapped inducing points. This has also been observed in regression tasks~\cite{bauer2016understanding}.}

\subsubsection{UCI benchmarks} Similar to~\cite{salimbeni2018orthogonally}, we conduct the comparison on 19 UCI datasets with $n \in [4897, 130064], d \in [7, 51]$, and $C \in [2, 26]$. The model configurations for comparison are detailed in Appendix~\ref{app_configurations}. Table~\ref{tab_UCI} summarizes the average classification accuracy (acc) and negative log likelihood (nll) results of scalable GPCs on the UCI benchmarks. The complete results on each dataset are provided in Tables~\ref{tab_UCI_acc} and~\ref{tab_UCI_nll} in Appendix~\ref{app_further_res}. Note that for the fully-connected neural networks using Selu activations, we use the results reported in~\cite{klambauer2017self}. For ORTH, we use the results reported in~\cite{salimbeni2018orthogonally}.\footnote{\cite{salimbeni2018orthogonally} employs the Mat\'{e}rn52+RBF kernel for ORTH. To be fair, the proposed GPCs use the same combination kernel and produce the results in Table~\ref{tab_GPCnoise} in Appendix~\ref{app_further_res}. It is found that the proposed scalable GPCs outperform ORTH.}

\begin{table*}[h]
	\caption{Average classification accuracy (acc) and negative log likelihood (nll) results on UCI benchmarks. The value of $\mathrm{GPCep}$ in the brackets indicates the average results for all the multi-class UCI datasets except \texttt{nursery}. The value of $\mathrm{GPCaug}$ in the brackets indicates the average results for all the multi-class UCI datasets except \texttt{waveform-noise}.}
	\centering
	\scalebox{1.}{
		\begin{tabular}{llcccccccc}%
			\hline
			& & Selu & ORTH & GPCep & GPCaug & GPC-I &  GPC-II & GPC-III & GPCsm \\
			\hline             
			binary & acc & 92.6 & 93.0 & 94.0 & 94.5 & \textbf{95.1} & \textbf{95.1} & \textbf{95.1} & NA \\
			& nll & NA & 0.1639 & 0.1506 & 0.1354 & 0.1355 & \textbf{0.1349} & \textbf{0.1349} & NA \\
			multi-class & acc & 91.1 & 89.2 & 87.1 (90.4) & 83.0 (87.5) & \textbf{93.0} & \textbf{93.0} & \textbf{93.0} & 92.6 \\
			& nll & NA & 0.5976 & 1.5637 (0.3168) & 0.3844 (0.3196) & 0.2374 & 0.2331 & 0.2353 & \textbf{0.1858} \\
			\hline
		\end{tabular}
	}
	\label{tab_UCI}
\end{table*}

It is observed that for binary benchmarks, the proposed three GPCs using different likelihoods outperform the others, and they are competitive in terms of average acc. Besides, compared to the GPC-I using step likelihood, GPC-II and GPC-III have a slightly better performance in terms of nll. As for multi-class benchmarks, all the proposed GPCs outperform the competitors, and particularly, the GPCsm using softmax likelihood provides remarkable performance in terms of nll. 

As for GPCep~\cite{hernandez2016scalable, villacampa2017scalable}, in order to have scalable and analytical ELBO, it employs (i) the FITC assumption $p(\mathbf{f}|\mathbf{u}) = \prod_{i=1}^n p(f_i|\mathbf{u})$, and (ii) an approximation to the integral in (6) of~\cite{villacampa2017scalable}. It is found that EP has a better approximation than VI for standard GPC~\cite{nickisch2008approximations}. The GPCep in our experiments indeed is very competitive, \textit{especially in terms of nll and convergence} (see for example Fig.~\ref{Fig_adult_waveform_noise_delta}(b)-(d)), in comparison to the proposed GPCs for almost all the datasets except \texttt{mushroom}, \texttt{nursery} and \texttt{wine-quality-white}. The capability of GPCep may be limited, because (i) the additional assumption and approximation may worsen the prediction~\cite{bauer2016understanding}; and (ii) the R package sometimes is unstable, e.g., it fails in 4 out of the 10 runs on the \texttt{mushroom} dataset.

The ORTH has acceptable acc results but provides relatively poor nll results in comparison to GPCep. Finally, the GPCaug exhibits competitive performance on the binary benchmarks. But the multi-class results of GPCaug are not attractive. For example, it fails on the \texttt{waveform-noise} dataset. Besides, the GPCaug sometimes suffers from slow convergence, see for example Fig.~\ref{Fig_adult_waveform_noise_delta}(c) and (d). This may be caused by the coordinate ascent optimizer deployed in the GPCaug package.

Finally, the proposed GPCs are much more efficient than GPCaug and GPCep due to the flexible tensorflow framework with GPU acceleration. For instance, the proposed GPCs require around 5 minutes to run on the \texttt{miniboone} dataset; while GPCep requires 7.9 hours and GPCaug requires 2.3 hours.

\subsubsection{The \texttt{airline} dataset} 
We finally assess the performance of the GPCs on the large-scale \texttt{airline} dataset containing the information of USA flights between January and April of 2008~\cite{hensman2015scalable}. The dataset has eight inputs including age of the aircraft, distance covered, airtime, departure time, arrival time, day of the week, day of the month and month. According to~\cite{villacampa2017scalable}, we treat the dataset as a classification task to predict the flight delay with three statuses: on time, more than 5 minutes of delay, or more than 5 minutes before time. We partition the dataset into 2M training points and 10000 test points. We employ the RBF kernel, and use $m=200$ for the proposed GPCs, GPCep and GPCaug, and $\beta=200$ and $\gamma=500$ for ORTH; we run the Adam optimizer over 100000 iterations using a mini-batch size of 1024 and the learning rate of 0.01. 

\begin{figure}[htb!]
	\centering
	\includegraphics[width=0.5\textwidth]{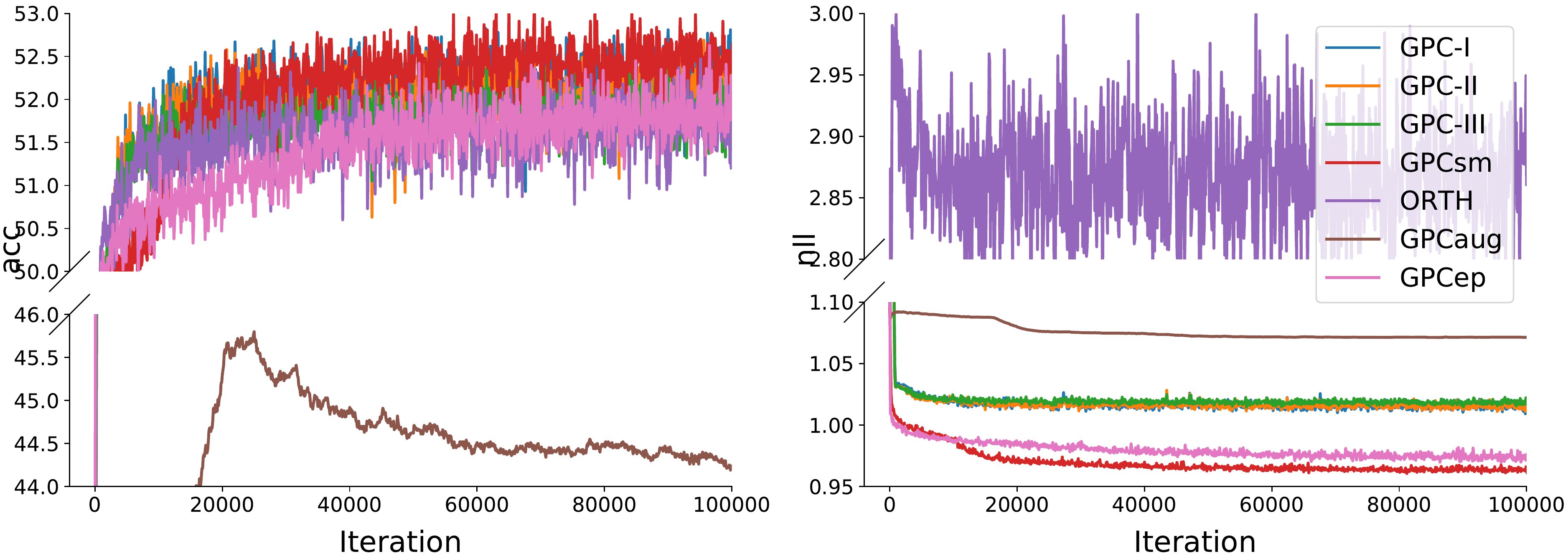}
	\caption{Comparative results of scalable GPCs on the large-scale \texttt{airline} dataset.}
	\label{fig_airline_comparison}
\end{figure}

As shown in Fig.~\ref{fig_airline_comparison}, the proposed GPCs, especially GPCsm, converge with superior results in terms of both acc and nll. The ORTH provides reasonable acc results, but suffers from poor nll results. In contrast, the GPCep converges with the second best nll results. Finally, the GPCaug has a stable but relatively poor performance on this dataset.

\subsection{Discussions of proposed GPCs} 
\subsubsection{Can we use a fixed $\delta$?}
The parameter $\delta$ is originally introduced in the robust step likelihoods~\eqref{eq_robust_step} and~\eqref{eq_multiclass_robust_step}. It prevents the logarithm in ELBOs~\eqref{eq_elbo_binary} and~\eqref{eq_elbo_multiclass} from being infinite; and plays the role of noise, like GPR, to be robust to outliers~\cite{kim2008outlier}. A fixed $\delta$, e.g., $10^{-3}$ is suggested by~\cite{de2016scalable, hensman2015mcmc} for classification. But we would like to argue that \textit{a fixed $\delta$ will worsen the predictive distribution, especially the predictive variance}.
	
\begin{figure}[htb!]
	\centering 
	\includegraphics[width=0.5\textwidth]{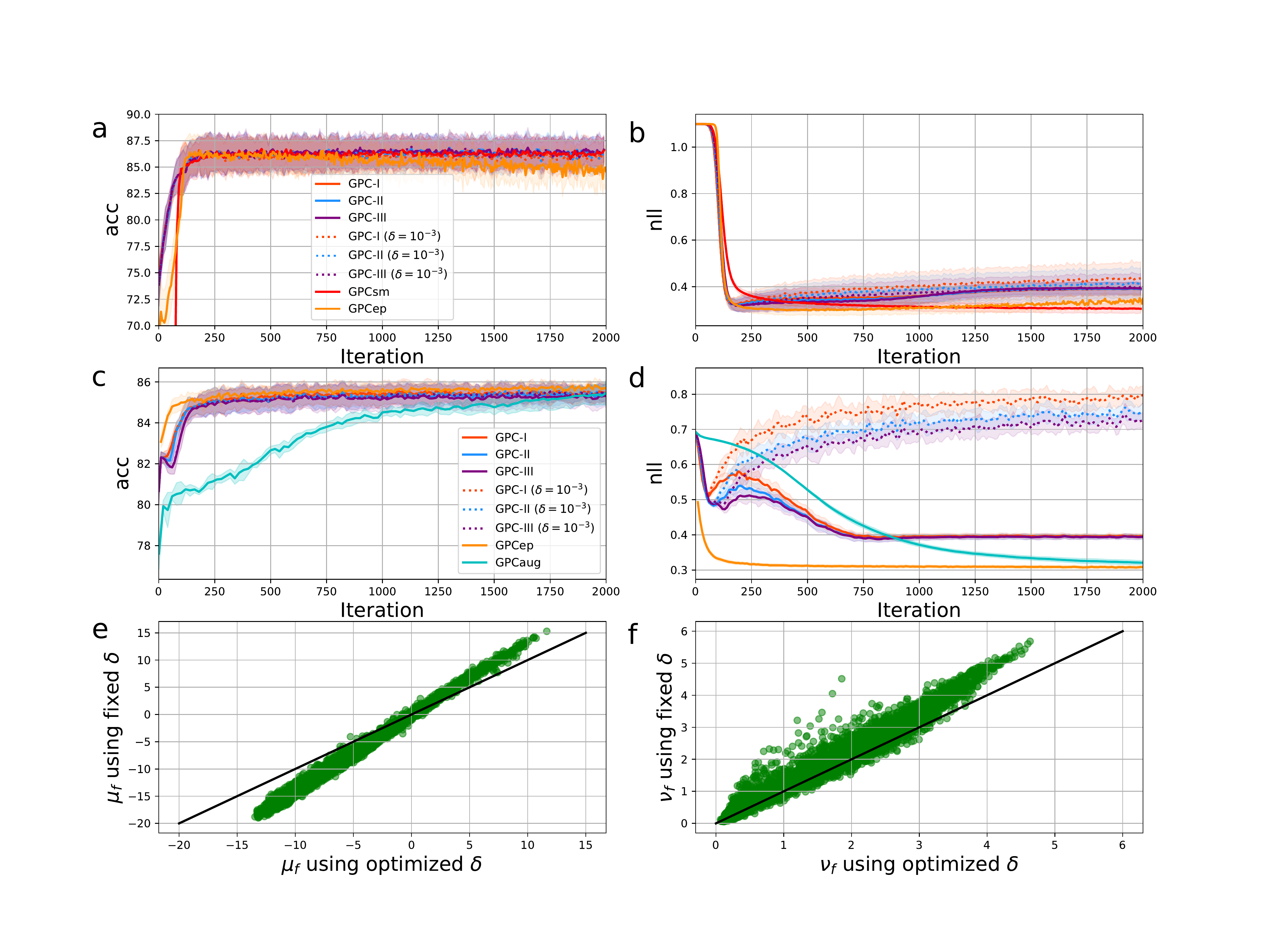}
	\caption{The accuracy (acc) and negative log likelihood (nll) of scalable GPCs on the multi-class \texttt{waveform-noise} dataset (top) and the binary \texttt{adult} dataset (medium); the bottom is the $\{\mu_{f_i}\}_{i=1}^n$ and $\{\nu_{f_i}\}_{i=1}^n$ estimated by GPC-I with fixed or optimized $\delta$ on the \texttt{adult} dataset. Note that the poor results of GPCaug for \texttt{waveform-noise} are not included.}
	\label{Fig_adult_waveform_noise_delta}
\end{figure}

To verify this, we run the proposed GPCs on the binary dataset \texttt{adult} and the multi-class dataset \texttt{waveform-noise} using a fixed $\delta$ and a free $\delta$, respectively, in Fig.~\ref{Fig_adult_waveform_noise_delta}(a)-(d). Particularly, Fig.~\ref{Fig_adult_waveform_noise_delta}(e) and (f) showcase the estimations of latent predictive mean $\{\mu_{f_i}\}_{i=1}^n$ and variance $\{\nu_{f_i}\}_{i=1}^n$ in GPC-I using fixed or optimized $\delta$ on the \texttt{adult} dataset.

We observe that the fixed $\delta$ brings a poor nll performance, e.g., it even increases during the training on the \texttt{adult} dataset. This phenomenon has also been observed in~\cite{villacampa2017scalable}. Taking the binary GPCs for example, we find that they have an optimum 
\begin{align}
\delta_* = 1 - \frac{1}{n}\sum_{i=1}^n \Phi_{\mathcal{N}}\left( \frac{y_i\mu_{f_i}}{\sqrt{a+\nu_{f_i}}} \right)
\end{align}
for the bound~\eqref{eq_elbo_binary}. When using a fixed $\delta$, the model has to \textit{over-/under-estimate} $\mu_{f_i}$ and $\nu_{f_i}$ in order to maximize $\mathcal{L}$, see Fig.~\ref{Fig_adult_waveform_noise_delta}(e) and (f). These ineffective estimations, especially the overestimated $\nu_{f_i}$, bring poor nll results. Hence, we treat $\delta > 0$ as a free hyperparameter and infer it from data in order to (i) relieve this issue, see Fig.~\ref{Fig_adult_waveform_noise_delta}(b) and (d); and (ii) keep the sum formulation in~\eqref{eq_elbo_binary}, which is crucial for stochastic optimization.

\subsubsection{Pros \& cons of proposed GPCs}
From the foregoing classification results, we have the following findings: 
\begin{itemize}
	\item It is found that the bounds~\eqref{eq_elbo_binary} and~\eqref{eq_elbo_multiclass} for the proposed GPCs using these three likelihoods (step, probit and logit) are expressed uniformly. Hence, the symmetric and bell-shaped error distributions help GPC-I, GPC-II and GPC-III perform similarly. Besides, increasing the noise variance $a$ in Gaussian error (step$\rightarrow$probit$\rightarrow$logit) brings softer likelihood and slightly better nll results; 
	\item The introduction of $\delta$ helps derive analytical ELBOs for GPC-I, GPC-II and GPC-III. And compared to the usage of a fixed $\delta$, optimizing $\delta$ significantly improves the nll results. However, it is found that the GPCs using $\delta$ have worse nll results than the counterparts on some binary/multi-class datasets, see for example Fig.~\ref{Fig_adult_waveform_noise_delta}; 
	\item Different from GPC-I, GPC-II and GPC-III, the asymmetric and varying Gumbel errors $\{\epsilon_i\}_{i=1}^n$ help GPCsm describe the \textit{heteroscedasticity}, thus outperforming the others in terms of nll. But the GPCsm may suffer from slow convergence in the early stage due to the complicated model structure, see for example Fig.~\ref{Fig_adult_waveform_noise_delta}(b).
\end{itemize}

\section{Conclusion}
\label{sec_conclusion}
This work presents a novel unifying framework which derives analytical ELBO for scalable GPCs using various likelihoods. This is achieved by introducing additive noises to interpret the GPCs in an augmented probability space through the internal variables $\mathbf{g}$ or directly the noises themselves. The superiority of our GPCs has been empirically demonstrated on extensive binary/multi-class classification tasks against state-of-the-art scalable GPCs.

\section*{Acknowledgments}
This work was conducted within the Rolls-Royce@NTU Corporate Lab with support from the National Research Foundation (NRF) Singapore under the Corp Lab@University Scheme. It is also partially supported by the Data Science and Artificial Intelligence Research Center (DSAIR) and the School of Computer Science and Engineering at Nanyang Technological University.

\appendices
\section{The Gaussian approximation to various error distributions}
\label{app_gaussian_approx}
\begin{figure}[htb!]
	\centering
	\includegraphics[width=0.5\textwidth]{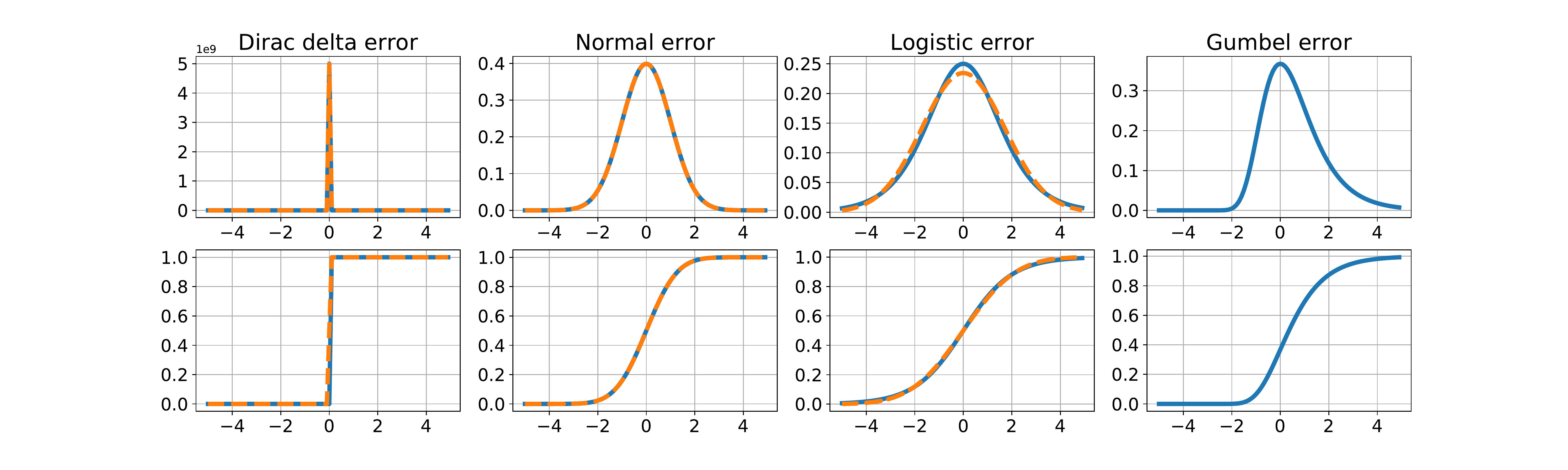}
	\caption{The error distributions related to various GPC likelihoods. The top row shows the PDF of error distributions, whereas the bottom row shows the CDF. The dash curves represent the Gaussian approximation to the error distributions.}
	\label{fig_errors}
\end{figure}

As discussed before, the step and (multinomial) probit/logit likelihoods can be recovered by varying over different error distributions in the GPC models~\eqref{eq_binary_GPC_noise} and~\eqref{eq_multiclass_GPC_noise}. Particularly, for the symmetric Dirac-delta, normal and logistic errors, since they are similar, we can describe them using a unifying Gaussian error $\mathcal{N}(\epsilon|0,a)$, resulting in a unifying GPC framework. As illustrated in Fig.~\ref{fig_errors}, the Dirac-delta error can be approximated by letting $a \rightarrow 0$; the logistic error can be approximated by having $a = 2.897$, which is derived by minimizing their maximal CDF difference. 

For the skewed Gumbel error, which relates to the specific softmax likelihood for multi-class GPC, we should however tackle it using another strategy, which is detailed in section~\ref{sec_GPCsm}.

\section{Treating the noise variance $a$ as a hyperparameter?}
\label{app_opt_a}
For the binary GPC with additive noise in section~\ref{sec_unifying_binary_GPC}, by treating the noise variance $a \ge 0$ as hyperparameters, we obtain the derivative of $\mathcal{L}$ in~\eqref{eq_elbo_binary} w.r.t $a$ as
\begin{align*}
\frac{\partial \mathcal{L}}{\partial a} = -\frac{1}{2} \log \left(\frac{1-\delta}{\delta} \right) \sum_{i=1}^{n} \left[ \phi_{\mathcal{N}} \left( \frac{y_i \mu_{f_i}}{\sqrt{a+\nu_{f_i}}} \right) \frac{y_i \mu_{f_i}}{(a+\nu_{f_i})^{\frac{3}{2}}}  \right].
\end{align*}
It is found that $\mu_{f_i}$ represents the prediction mean at $\mathbf{x}_i$. When the binary GPC provides sensible predictions, we have $y_i \mu_{f_i} > 0$. Hence, we have $\partial \mathcal{L} / \partial a < 0$, which means that the ELBO $\mathcal{L}$ arrives at the maximum when $a = 0$.

Similarly, for the multi-class GPC with additive noise in section~\ref{sec_unifying_multiclass_GPC}, we obtain the derivative of $\mathcal{L}$ in~\eqref{eq_elbo_multiclass} w.r.t $a$ as
\begin{equation*}
\frac{\partial \mathcal{L}}{\partial a} = \log \left(\frac{1-\delta}{\delta} (C-1) \right) \sum_{i=1}^n \frac{\partial S_i}{\partial a},
\end{equation*}
where
\begin{align*}
\frac{\partial S_i}{\partial a} =& -\frac{1}{2} \mathbb{E}_{g_i^{y_i}} \left[ \sum_{c \ne y_i} \frac{g_i^{y_i}-\mu_{f_i}^c}{(a + \nu_{f_i}^c)^{\frac{3}{2}}} \phi_{\mathcal{N}} \left(\frac{g_i^{y_i}-\mu_{f_i}^c}{\sqrt{a + \nu_{f_i}^c}} \right) \right. \\
& \left. \prod_{c' \ne y_i, c} \Phi_{\mathcal{N}} \left(\frac{g_i^{y_i}-\mu_{f_i}^{c'}}{\sqrt{a + \nu_{f_i}^{c'}}} \right) \right] \\
&+ \int \frac{\partial \phi_{\mathcal{N}} \left(\frac{g_i^{y_i}-\mu_{f_i}^{y_i}}{\sqrt{a + \nu_{f_i}^{y_i}}} \right)}{\partial a} \prod_{c \ne y_i} \Phi_{\mathcal{N}} \left(\frac{g_i^{y_i}-\mu_{f_i}^c}{\sqrt{a + \nu_{f_i}^c}} \right) dg_i^{y_i}.
\end{align*}
For the first term in the right-hand side of $\partial S_i / \partial a$, since $\Phi_{\mathcal{N}}(x) > \Phi_{\mathcal{N}}(-x)$ and $\phi_{\mathcal{N}}(x) = \phi_{\mathcal{N}}(-x)$ for $x > 0$, we know that this is a negative term. For the second term in the right-hand side of $\partial S_i / \partial a$, it satisfies
\begin{align*}
&< \int \frac{\partial \phi_{\mathcal{N}} \left(\frac{g_i^{y_i}-\mu_{f_i}^{y_i}}{\sqrt{a + \nu_{f_i}^{y_i}}} \right)}{\partial a} dg_i^{y_i} \\
&= -\frac{1}{2} \mathbb{E}_{g_i^{y_i}} \left[ \frac{1}{a + \nu_{f_i}^{y_i}} - \frac{(g_i^{y_i}-\mu_{f_i}^{y_i})^2}{(a + \nu_{f_i}^{y_i})^2} \right] = 0.
\end{align*}
Hence, we again have $\partial S_i / \partial a < 0$ and furthermore $\partial \mathcal{L} / \partial a < 0$, which means that the ELBO $\mathcal{L}$ arrives at the maximum when $a = 0$.

The above analysis reveals that $a$ cannot play a role of hyperparameter.

\section{The ELBO for multi-class GPC using softmax likelihood}
\label{app_elbo_softmax}
The original ELBO of multi-class GPC using softmax likelihood in section~\ref{sec_GPCsm} is
\begin{align*}
\mathcal{L} =& \sum_i^n \left\langle \log p(y_i| \mathbf{f}_i, \epsilon_i) \right\rangle_{q(\epsilon_i|\mathbf{f}_i)q(\mathbf{f}_i)} - \sum_{i=1}^n\mathrm{KL}(q(\epsilon_i|\mathbf{f}_i) || p(\epsilon_i)) \\
&-  \sum_{c=1}^C\mathrm{KL}(q(\mathbf{u}^c|\mathbf{y}) || p(\mathbf{u}^c)).
\end{align*}
For the first double-expectation term in the right-hand side of $\mathcal{L}$, we first calculate the inner expectation as
\begin{equation*}
\begin{aligned}
&\left\langle \log p(y_i| \mathbf{f}_i, \epsilon_i) \right\rangle_{q(\epsilon_i|\mathbf{f}_i)} \\
=& \sum_{c \ne y_i} \int_{-\infty}^{+\infty} q(\epsilon_i|\mathbf{f}_i) \log \Phi_{\mathcal{G}}(\epsilon_i+f_i^{y_i}-f_i^c) d\epsilon_i \\
=& -\sum_{c \ne y_i} \int_{-\infty}^{+\infty} e^{\left(-(\epsilon_i-\log\theta_i)-e^{-(\epsilon_i-\log\theta_i)} \right)} e^{-(\epsilon_i+f_i^{y_i}-f_i^c)} d\epsilon_i \\
\overset{u=e^{-\epsilon_i}}{=}& -\sum_{c \ne y_i} \theta_i e^{f_i^c-f_i^{y_i}} \int_{0}^{+\infty} u e^{-\theta u}du \\
=& \sum_{c \ne y_i} \theta_i e^{f_i^c-f_i^{y_i}} \left. \frac{(1+\theta u)e^{-\theta u}}{\theta^2}\right|_{0}^{+\infty} \\
=& -\frac{1}{\theta_i} \sum_{c \ne y_i} e^{f_i^c-f_i^{y_i}}.
\end{aligned}
\end{equation*}
Then, the outside expectation is analytically expressed as
\begin{equation*}
\begin{aligned}
&\left\langle \log p(y_i| \mathbf{f}_i, \epsilon_i) \right\rangle_{q(\epsilon_i|\mathbf{f}_i) q(\mathbf{f}_i)} \\
=& -\frac{1}{\theta_i} \int e^{-f_i^{y_i}} q(f_i^{y_i}) df_i^{y_i} \sum_{c \ne y_i} \int e^{f_i^c} q(f_i^c) df_i^c \\
=& -\frac{1}{\theta_i} \exp \left(\frac{\nu_{f_i}^c}{2}-\mu_{f_i}^c\right) \sum_{c \ne y_i} \exp \left(\frac{\nu_{f_i}^c}{2}+\mu_{f_i}^c\right).
\end{aligned}
\end{equation*}
For the KL divergence $\sum_{i=1}^n\mathrm{KL}(q(\epsilon_i|\mathbf{f}_i) || p(\epsilon_i))$ in the right-hand side of $\mathcal{L}$, we have
\begin{equation*}
\sum_{i=1}^n\mathrm{KL}(q(\epsilon_i|\mathbf{f}_i) || p(\epsilon_i)) = \sum_{i=1}^n \left(\log\theta_i+\frac{1}{\theta_i}-1\right).
\end{equation*}
For the KL divergence $\sum_{c=1}^C \mathrm{KL}(q(\mathbf{u}^c|\mathbf{y}) || p(\mathbf{u}^c))$ we have
\begin{equation*}
\begin{aligned}
:=& \frac{1}{2} \sum_{c=1}^C  \left( \log \frac{|\mathbf{K}_{mm}^c|}{|\mathbf{S}^c|} - m + \mathrm{tr}[(\mathbf{K}_{mm}^c)^{-1} \mathbf{S}^c] \right. \\
&\left. + (\mathbf{m}^c)^{\mathsf{T}} (\mathbf{K}_{mm}^c)^{-1} \mathbf{m}^c  \right).
\end{aligned}
\end{equation*}
According to the above computations, the ELBO is reorganized as
\begin{equation*}
\mathcal{L} = \sum_{i=1}^n \left\{
-\frac{1}{\theta_i} \mathcal{P}_i - \log\theta_i-\frac{1}{\theta_i}+1 
\right\} -\sum_{c=1}^C\mathrm{KL}(q(\mathbf{u}^c|\mathbf{y}) || p(\mathbf{u}^c)),
\end{equation*}
with $\mathcal{P}_i = \exp \left(\frac{\nu_{f_i}^{y_i}}{2}-\mu_{f_i}^{y_i}\right) \sum_{c \ne y_i} \exp \left(\frac{\nu_{f_i}^c}{2}+\mu_{f_i}^c\right)$. The double-sum in the right-hand side of $\mathcal{L}$ over data points and classes implies that we could obtain an unbiased estimation of $\mathcal{L}$ via a subset of both data points and classes, which is potential for large categorical cases~\cite{ruiz2018augment}.

\section{Details for UCI experiments}
\label{app_configurations}

For the experiments on 19 UCI classification benchmarks, similar to~\cite{salimbeni2018orthogonally}, we employ the experimental settings detailed as follows.

As for data preprocessing, we use the package \texttt{Bayesian Benchmarks}\footnote{\url{https://github.com/hughsalimbeni/bayesian_benchmarks}.} which normalizes the inputs of datasets to have zero mean and unit variance along each dimension, and randomly chooses 10\% of the data as the test set.

As for the kernel function, the results of ORTH in~\cite{salimbeni2018orthogonally} employ a combination of the Mat\'{e}rn52 kernel with length-scale of $0.1 \sqrt{d}$ and variance of 5.0, and the RBF kernel with length-scale of $\sqrt{d}$ and variance of 5.0. But for GPCep and GPCaug, since their packages either only support the RBF kernel or have no kernel combination module, we adopt the RBF kernel with length-scale of $0.1 \sqrt{d}$ and variance of 5.0 in the comparison. The proposed GPCs also use the RBF kernel in Tables~\ref{tab_UCI_acc} and~\ref{tab_UCI_nll} in Appendix~\ref{app_further_res}. Additionally, to compare with ORTH, Table~\ref{tab_GPCnoise} in Appendix~\ref{app_further_res} offers the results of proposed GPCs using the Mat\'{e}rn52+RBF kernel. 

As for optimization, we use $m=300$ inducing points initialized through the \textit{k}-means technique, and a mini-batch size of 1024. We employ the Adam optimizer for all the scalable GPCs except GPCaug,\footnote{The Julia package of GPCaug employs a coordinate ascent optimizer.} and run it over 10000 iterations with a learning rate of 0.01. We here do not employ the natural gradient descent (NGD) strategy for optimizing the variational parameters. Because in comparison to the regression task, the NGD+Adam optimizer brings little benefits for classification~\cite{salimbeni2018natural,salimbeni2018orthogonally}. 

Finally, as for likelihoods, the ORTH and GPCep employ the probit likelihood for binary case and the step likelihood for multi-class case; the GPCaug adopts the logit likelihood for binary case and the logistic-softmax likelihood for multi-class case; and our GPCs have no limit and run with all likelihoods.

\section{Detailed results}
\label{app_further_res}
Tables~\ref{tab_UCI_acc} and~\ref{tab_UCI_nll} provide the average results (over 10 runs) of scalable binary/multi-class GPCs on 19 UCI benchmarks in terms of accuracy and negative log likelihood. Additionally, Table~\ref{tab_GPCnoise} shows the average results of proposed GPCs using the Mat\'{e}rn52+RBF kernel on 19 UCI benchmarks.

\begin{table*}
	\caption{Average classification accuracy (acc) results on 19 UCI benchmarks. Note that all the GPCs except ORTH use the RBF kernel in this comparison. The results of ORTH reported in~\cite{salimbeni2018orthogonally} employ the Mat\'{e}rn52+RBF kernel.} 
	\label{tab_UCI_acc}
	\centering
	\resizebox{0.8\textwidth}{!}{%
		\begin{tabular}{llllllllllll}
			\hline
			&$n$ &$d$ &$C$ &Selu & ORTH & GPCep & GPCaug & GPC-I &    GPC-II & GPC-III &  GPCsm \\
			\hline
			adult	&48842	&15	&2	&84.67	&85.65	&85.64	&85.60	&85.53 	&85.47 	&85.55	&NA	\\
			connect-4	&67557	&43	&2	&88.07	&85.99	&84.40	&81.28	&84.86 	&84.68 	&84.52 &NA	\\
			magic	&19020	&11	&2	&86.92	&89.35	&99.89	&99.64	&99.89 	&99.92 	&99.90 &NA	\\
			miniboone	&130064	&51	&2	&93.07	&93.49	&99.77	&99.87	&99.89 	&99.92 	&99.92 &NA	\\
			mushroom	&8124	&22	&2	&100.00	&100.00	&92.03	&100.00	&99.98 	&99.98 	&99.98 &NA	\\
			ringnorm	&7400	&21	&2	&97.51	&98.78	&98.43	&97.62	&98.22 	&98.35 	&98.34 &NA	\\
			twonorm	&7400	&21	&2	&98.05	&97.65	&97.78	&97.78	&97.45 	&97.49 	&97.58 &NA	\\
			chess-krvk	&28056	&7	&18	&88.05	&67.76	&99.89	&88.39	&99.82 	&99.78 	&99.72 	&99.85 \\
			letter	&20000	&17	&26	&97.26	&95.77	&95.43	&83.76	&96.50 	&96.66 	&96.49 	&96.25 \\
			nursery	&12960	&9	&5	&99.78	&97.30	&50.64	&92.67	&99.99 	&99.98 	&99.99 	&99.63 \\
			page-blocks	&5473	&11	&5	&95.83	&97.21	&97.41	&96.02	&97.06 	&97.26 	&97.03 	&97.10 \\
			pendigits	&10992	&17	&10	&97.06	&99.66	&99.36	&98.53	&99.50 	&99.58 	&99.59 	&99.45 \\
			statlog-landsat	&6435	&37	&6	&91.00	&91.28	&92.55	&85.68	&93.37 	&93.76 	&93.70 	&91.91 \\
			statlog-shuttle	&58000	&10	&7	&99.90	&99.90	&99.92	&99.70	&99.93 	&99.91 	&99.90 	&99.88 \\
			thyroid	&7200	&22	&3	&98.16	&99.47	&99.33	&94.83	&99.29 	&99.39 	&99.39 	&99.21 \\
			wall-following	&5456	&25	&4	&90.98	&95.56	&96.78	&88.42	&96.96 	&96.70 	&96.72 	&96.50 \\
			waveform	&5000	&22	&3	&84.80	&86.13	&84.96	&86.52	&86.86 	&87.02 	&87.34 	&87.84 \\
			waveform-noise	&5000	&41	&3	&86.08	&82.93	&85.88	&33.40	&85.16 	&85.30 	&85.54 	&86.24 \\
			wine-quality-white	&4898	&12	&7	&63.73	&57.05	&42.90 &48.49	&61.22 	&60.65 	&60.51 	&57.24 \\    
			\hline
		\end{tabular}
	}
\end{table*}

\begin{table*}
	\caption{Average negative log likelihood (nll) results on 19 UCI benchmarks. Note that all the GPCs except ORTH use the RBF kernel in this comparison. The results of ORTH reported in~\cite{salimbeni2018orthogonally} employ the Mat\'{e}rn52+RBF kernel.} 
	\label{tab_UCI_nll}
	\centering
	\resizebox{0.8\textwidth}{!}{%
		\begin{tabular}{lllllllllll}
			\hline
			&$n$ &$d$ &$C$ & ORTH & GPCep & GPCaug & GPC-I &    GPC-II & GPC-III &  GPCsm \\
			\hline
			adult	&48842	&15	&2	&0.3045	&0.3074	&0.3306	&0.3971 	&0.3956 	&0.3933 &NA	\\
			connect-4	&67557	&43	&2	&0.3086	&0.3822	&0.4534	&0.3958 	&0.3998 	&0.4029 &NA	\\
			magic	&19020	&11	&2	&0.2658	&0.0039	&0.0153	&0.0041 	&0.0030 	&0.0029 &NA	\\
			miniboone	&130064	&51	&2	&0.1618	&0.0031	&0.0064	&0.0046 	&0.0028 	&0.0029 &NA	\\
			mushroom	&8124	&22	&2	&0.0009	&0.2363	&0.0054	&0.0004 	&0.0004 	&0.0004 &NA	\\
			ringnorm &7400	&21	&2	&0.0466	&0.0507	&0.0739	&0.0568 	&0.0593 	&0.0619 &NA	\\
			twonorm	&7400	&21	&2	&0.0590	&0.0707	&0.0625	&0.0898 	&0.0831 	&0.0802 &NA	\\
			chess-krvk	&28056	&7	&18	&2.1625	&0.0236	&0.1918	&0.0141 	&0.0169 	&0.0223 	&0.0072 \\
			letter	&20000	&17	&26	&0.2276	&0.1934	&0.6852	&0.1685 	&0.1699 	&0.1789 	&0.1528 \\
			nursery	&12960	&9	&5	&0.2225	&15.2799	&0.1546	&0.0042 	&0.0074 	&0.0070 	&0.0299 \\
			page-blocks	&5473	&11	&5	&0.1328	&0.0763	&0.1343	&0.1314 	&0.1165 	&0.1226 	&0.0896 \\
			pendigits	&10992	&17	&10	&0.0209	&0.0258	&0.1095	&0.0225 	&0.0223 	&0.0232 	&0.0282 \\
			statlog-landsat	&6435	&37	&6	&0.3956	&0.2148	&0.3356	&0.2280 	&0.2164 	&0.2195 	&0.2110 \\
			statlog-shuttle	&58000	&10	&7	&0.0049	&0.0035	&0.0179	&0.0036 	&0.0041 	&0.0047 	&0.0045 \\
			thyroid	&7200	&22	&3	&0.0115	&0.0205	&0.0893	&0.0296 	&0.0285 	&0.0293 	&0.0245 \\
			wall-following	&5456	&25	&4	&0.1514	&0.0839	&0.3437	&0.1124 	&0.1187 	&0.1242 	&0.1033 \\
			waveform	&5000	&22	&3	&0.5640	&0.4641	&0.3342	&0.4097 	&0.3987 	&0.3948 	&0.2832 \\
			waveform-noise	&5000	&41	&3	&0.7096	&0.3170	&1.0980	&0.4437 	&0.4295 	&0.4223 	&0.2999 \\
			wine-quality-white	&4898	&12	&7	&2.5681	&2.0612	&1.1191	&1.2806 	&1.2684 	&1.2748 	&0.9953 \\
			\hline
		\end{tabular}
	}
\end{table*}

\begin{table*}
	\caption{Average results of proposed GPCs on 19 UCI benchmarks using the Mat\'{e}rn52+RBF kernel. Note that the proposed GPCs provide relatively poor results on the \texttt{connect-4} dataset, which skew the average performance.} 
	\label{tab_GPCnoise}
	\centering
	\resizebox{0.8\textwidth}{!}{%
		\begin{tabular}{llll|llll|llll}
			\hline
			& & & & \multicolumn{4}{c|}{acc} & \multicolumn{4}{c}{nll} \\
			&$n$ &$d$ &$C$ & GPC-I & GPC-II & GPC-III &  GPCsm & GPC-I &    GPC-II & GPC-III &  GPCsm \\
			\hline
			adult	&48842	&15	&2	&85.75	&85.86	&85.76	&NA &0.3921	&0.3892	&0.3900	&NA	\\
			connect-4	&67557	&43	&2	&76.43	&76.49	&76.38 &NA	&0.5393	&0.5383	&0.5396	&NA	\\
			magic	&19020	&11	&2	&99.89	&99.94	&99.95 &NA	&0.0038	&0.0029	&0.0024	&NA	\\
			miniboone	&130064	&51	&2	&99.87	&99.93	&99.92 &NA	&0.0049	&0.0028	&0.0030	&NA	\\
			mushroom	&8124	&22	&2	&100.00	&100.00	&100.00 &NA	&0.0001	&0.0001	&0.0002	&NA	\\
			ringnorm &7400	&21	&2	&98.23	&98.32	&98.35 &NA	&0.0568	&0.0597	&0.0624	&NA	\\
			twonorm	&7400	&21	&2	&97.45	&97.49	&97.64 &NA	&0.0932	&0.0830	&0.0798	&NA	\\
			chess-krvk	&28056	&7	&18	&99.67	&99.66	&99.64 &99.80	&0.0277	&0.0275	&0.0290	&0.0076	\\
			letter	&20000	&17	&26	&95.91	&96.28	&96.26 &96.36	&0.2101	&0.2008	&0.2041	&0.1433	\\
			nursery	&12960	&9	&5	&97.54	&97.54	&97.54 &99.95	&0.1500	&0.1502	&0.1504	&0.0120	\\
			page-blocks	&5473	&11	&5	&97.83	&97.79	&97.85 &97.66	&0.0990	&0.0934	&0.0903	&0.0745	\\
			pendigits	&10992	&17	&10	&99.55	&99.60	&99.62 &99.45	&0.0226	&0.0211	&0.0218	&0.0255	\\
			statlog-landsat	&6435	&37	&6	&93.65	&93.90	&93.77 &92.80	&0.2189	&0.2043	&0.2037	&0.1900	\\
			statlog-shuttle	&58000	&10	&7	&99.94	&99.94	&99.94 &99.93	&0.0028	&0.0030	&0.0033	&0.0029	\\
			thyroid	&7200	&22	&3	&99.25	&99.42	&99.38 &99.46	&0.0269	&0.0236	&0.0231	&0.0178	\\
			wall-following	&5456	&25	&4	&97.82	&97.78	&97.75 &97.66	&0.0727	&0.0733	&0.0763	&0.0694	\\
			waveform	&5000	&22	&3	&86.96	&86.84	&86.96 &88.06	&0.4130	&0.4058	&0.4000	&0.2828	\\
			waveform-noise	&5000	&41	&3	&84.86	&85.36	&85.66 &86.22	&0.4551	&0.4280	&0.4224	&0.2995	\\
			wine-quality-white	&4898	&12	&7	&60.71	&60.51	&60.96 &58.20	&1.2662	&1.2623	&1.2573	&0.9868	\\
			\hline
			\multicolumn{4}{c|}{binary (average)} & 93.9 & 94.0 & 94.0 & NA & 0.1557 & 0.1537 & 0.1539 & NA \\
			\multicolumn{4}{c|}{multi-class (average)} & 92.8 & 92.9 & 92.9 & 93.0 & 0.2471 & 0.2411 & 0.2401 & 0.1760 \\
			\hline
		\end{tabular}
	}
\end{table*}

\ifCLASSOPTIONcaptionsoff
  \newpage
\fi



%

\bibliographystyle{IEEEtran}
\bibliography{IEEEabrv,GPCnoise}

\begin{thebibliography}{10}
\providecommand{\url}[1]{#1}
\csname url@samestyle\endcsname
\providecommand{\newblock}{\relax}
\providecommand{\bibinfo}[2]{#2}
\providecommand{\BIBentrySTDinterwordspacing}{\spaceskip=0pt\relax}
\providecommand{\BIBentryALTinterwordstretchfactor}{4}
\providecommand{\BIBentryALTinterwordspacing}{\spaceskip=\fontdimen2\font plus
\BIBentryALTinterwordstretchfactor\fontdimen3\font minus
  \fontdimen4\font\relax}
\providecommand{\BIBforeignlanguage}[2]{{%
\expandafter\ifx\csname l@#1\endcsname\relax
\typeout{** WARNING: IEEEtran.bst: No hyphenation pattern has been}%
\typeout{** loaded for the language `#1'. Using the pattern for}%
\typeout{** the default language instead.}%
\else
\language=\csname l@#1\endcsname
\fi
#2}}
\providecommand{\BIBdecl}{\relax}
\BIBdecl

\bibitem{rasmussen2006gaussian}
C.~E. Rasmussen and C.~K. Williams, \emph{Gaussian processes for machine
  learning}.\hskip 1em plus 0.5em minus 0.4em\relax MIT Press, 2006.

\bibitem{settles1994active}
B.~Settles, ``Active learning literature survey,'' \emph{Machine Learning},
  vol.~15, no.~2, pp. 201--221, 1994.

\bibitem{lawrence2005probabilistic}
N.~Lawrence, ``Probabilistic non-linear principal component analysis with
  {G}aussian process latent variable models,'' \emph{Journal of Machine
  Learning Research}, vol.~6, no. Nov, pp. 1783--1816, 2005.

\bibitem{alvarez2012kernels}
M.~A. Alvarez, L.~Rosasco, N.~D. Lawrence \emph{et~al.}, ``Kernels for
  vector-valued functions: {A} review,'' \emph{Foundations and
  Trends{\textregistered} in Machine Learning}, vol.~4, no.~3, pp. 195--266,
  2012.

\bibitem{liu2018remarks}
H.~Liu, J.~Cai, and Y.-S. Ong, ``Remarks on multi-output {G}aussian process
  regression,'' \emph{Knowledge-Based Systems}, vol. 144, no. March, pp.
  102--121, 2018.

\bibitem{kim2006bayesian}
H.-C. Kim and Z.~Ghahramani, ``Bayesian {G}aussian process classification with
  the {EM-EP} algorithm,'' \emph{IEEE Transactions on Pattern Analysis and
  Machine Intelligence}, vol.~28, no.~12, pp. 1948--1959, 2006.

\bibitem{wang2013spectrum}
L.~Wang and C.~Li, ``Spectrum-based kernel length estimation for {G}aussian
  process classification,'' \emph{IEEE Transactions on Cybernetics}, vol.~44,
  no.~6, pp. 805--816, 2013.

\bibitem{quinonero2005unifying}
J.~Qui{\~n}onero-Candela and C.~E. Rasmussen, ``A unifying view of sparse
  approximate {G}aussian process regression,'' \emph{Journal of Machine
  Learning Research}, vol.~6, no. Dec, pp. 1939--1959, 2005.

\bibitem{titsias2009variational}
M.~K. Titsias, ``Variational learning of inducing variables in sparse
  {G}aussian processes,'' in \emph{Artificial Intelligence and Statistics},
  2009, pp. 567--574.

\bibitem{hensman2013gaussian}
J.~Hensman, N.~Fusi, and N.~D. Lawrence, ``Gaussian processes for big data,''
  in \emph{Uncertainty in Artificial Intelligence}.\hskip 1em plus 0.5em minus
  0.4em\relax Citeseer, 2013, pp. 282--290.

\bibitem{frohlich2013large}
B.~Fr{\"o}hlich, E.~Rodner, M.~Kemmler, and J.~Denzler, ``Large-scale
  {G}aussian process multi-class classification for semantic segmentation and
  facade recognition,'' \emph{Machine Vision and Applications}, vol.~24, no.~5,
  pp. 1043--1053, 2013.

\bibitem{milios2018dirichlet}
D.~Milios, R.~Camoriano, P.~Michiardi, L.~Rosasco, and M.~Filippone,
  ``Dirichlet-based {G}aussian processes for large-scale calibrated
  classification,'' in \emph{Advances in Neural Information Processing
  Systems}, 2018, pp. 6008--6018.

\bibitem{nickisch2008approximations}
H.~Nickisch and C.~E. Rasmussen, ``Approximations for binary {G}aussian process
  classification,'' \emph{Journal of Machine Learning Research}, vol.~9, no.
  Oct, pp. 2035--2078, 2008.

\bibitem{liu2018gaussian}
H.~Liu, Y.-S. Ong, X.~Shen, and J.~Cai, ``When {G}aussian process meets big
  data: {A} review of scalable gps,'' \emph{arXiv preprint arXiv:1807.01065},
  2018.

\bibitem{snelson2006sparse}
E.~Snelson and Z.~Ghahramani, ``Sparse {G}aussian processes using
  pseudo-inputs,'' in \emph{Advances in Neural Information Processing Systems},
  2006, pp. 1257--1264.

\bibitem{wilson2015kernel}
A.~Wilson and H.~Nickisch, ``Kernel interpolation for scalable structured
  {G}aussian processes ({KISS-GP}),'' in \emph{International Conference on
  Machine Learning}, 2015, pp. 1775--1784.

\bibitem{hoang2015unifying}
T.~N. Hoang, Q.~M. Hoang, and B.~K.~H. Low, ``A unifying framework of anytime
  sparse {G}aussian process regression models with stochastic variational
  inference for big data,'' in \emph{International Conference on Machine
  Learning}, 2015, pp. 569--578.

\bibitem{peng2017asynchronous}
H.~Peng, S.~Zhe, X.~Zhang, and Y.~Qi, ``Asynchronous distributed variational
  {G}aussian process for regression,'' in \emph{International Conference on
  Machine Learning}, 2017, pp. 2788--2797.

\bibitem{hoffman2013stochastic}
M.~D. Hoffman, D.~M. Blei, C.~Wang, and J.~Paisley, ``Stochastic variational
  inference,'' \emph{Journal of Machine Learning Research}, vol.~14, no.~1, pp.
  1303--1347, 2013.

\bibitem{pleiss2018constant}
G.~Pleiss, J.~Gardner, K.~Weinberger, and A.~G. Wilson, ``Constant-time
  predictive distributions for {G}aussian processes,'' in \emph{International
  Conference on Machine Learning}, 2018, pp. 4111--4120.

\bibitem{gardner2018product}
J.~Gardner, G.~Pleiss, R.~Wu, K.~Weinberger, and A.~Wilson, ``Product kernel
  interpolation for scalable {G}aussian processes,'' in \emph{Artificial
  Intelligence and Statistics}, 2018, pp. 1407--1416.

\bibitem{naish2008generalized}
A.~Naish-Guzman and S.~Holden, ``The generalized {FITC} approximation,'' in
  \emph{Advances in Neural Information Processing Systems}, 2008, pp.
  1057--1064.

\bibitem{bauer2016understanding}
M.~Bauer, M.~van~der Wilk, and C.~E. Rasmussen, ``Understanding probabilistic
  sparse {G}aussian process approximations,'' in \emph{Advances in Neural
  Information Processing Systems}, 2016, pp. 1533--1541.

\bibitem{hernandez2016scalable}
D.~Hern{\'a}ndez-Lobato and J.~M. Hern{\'a}ndez-Lobato, ``Scalable {G}aussian
  process classification via expectation propagation,'' in \emph{Artificial
  Intelligence and Statistics}, 2016, pp. 168--176.

\bibitem{villacampa2017scalable}
C.~Villacampa-Calvo and D.~Hern{\'a}ndez-Lobato, ``Scalable multi-class
  {G}aussian process classification using expectation propagation,'' in
  \emph{International Conference on Machine Learning}.\hskip 1em plus 0.5em
  minus 0.4em\relax JMLR. org, 2017, pp. 3550--3559.

\bibitem{li2015stochastic}
Y.~Li, J.~M. Hern{\'a}ndez-Lobato, and R.~E. Turner, ``Stochastic expectation
  propagation,'' in \emph{Advances in Neural Information Processing Systems},
  2015, pp. 2323--2331.

\bibitem{hensman2015scalable}
J.~Hensman, A.~Matthews, and Z.~Ghahramani, ``Scalable variational gaussian
  process classification,'' in \emph{Artificial Intelligence and Statistics},
  2015, pp. 351--360.

\bibitem{hensman2015mcmc}
J.~Hensman, A.~G. Matthews, M.~Filippone, and Z.~Ghahramani, ``{MCMC} for
  variationally sparse {G}aussian processes,'' in \emph{Advances in Neural
  Information Processing Systems}, 2015, pp. 1648--1656.

\bibitem{polson2013bayesian}
N.~G. Polson, J.~G. Scott, and J.~Windle, ``Bayesian inference for logistic
  models using p{\'o}lya--{G}amma latent variables,'' \emph{Journal of the
  American Statistical Association}, vol. 108, no. 504, pp. 1339--1349, 2013.

\bibitem{wenzel2018efficient}
F.~Wenzel, T.~Galy-Fajou, C.~Donner, M.~Kloft, and M.~Opper, ``Efficient
  {G}aussian process classification using p{\`o}lya-gamma data augmentation,''
  in \emph{International Conference on Machine Learning}, 2018.

\bibitem{galy2019multi}
T.~Galy-Fajou, F.~Wenzel, C.~Donner, and M.~Opper, ``Multi-class gaussian
  process classification made conjugate: Efficient inference via data
  augmentation,'' in \emph{International Conference on Artificial Intelligence
  and Statistics}, 2019.

\bibitem{cheng2017variational}
C.-A. Cheng and B.~Boots, ``Variational inference for {G}aussian process models
  with linear complexity,'' in \emph{Advances in Neural Information Processing
  Systems}, 2017, pp. 5184--5194.

\bibitem{salimbeni2018orthogonally}
H.~Salimbeni, C.-A. Cheng, B.~Boots, and M.~Deisenroth, ``Orthogonally
  decoupled variational {G}aussian processes,'' in \emph{Advances in Neural
  Information Processing Systems}, 2018, pp. 8711--8720.

\bibitem{matthews2017gpflow}
D.~G. Matthews, G.~Alexander, M.~Van Der~Wilk, T.~Nickson, K.~Fujii,
  A.~Boukouvalas, P.~Le{\'o}n-Villagr{\'a}, Z.~Ghahramani, and J.~Hensman,
  ``{GP}flow: {A} {G}aussian process library using tensorflow,'' \emph{Journal
  of Machine Learning Research}, vol.~18, no.~1, pp. 1299--1304, 2017.

\bibitem{bowling2009logistic}
S.~R. Bowling, M.~T. Khasawneh, S.~Kaewkuekool, and B.~R. Cho, ``A logistic
  approximation to the cumulative normal distribution,'' \emph{Journal of
  Industrial Engineering and Management}, vol.~2, no.~1, 2009.

\bibitem{de2016scalable}
A.~G. de~Garis~Matthews, ``Scalable {G}aussian process inference using
  variational methods,'' \emph{Department of Engineering, University of
  Cambridge}, 2016.

\bibitem{kingma2014adam}
D.~P. Kingma and J.~Ba, ``Adam: {A} method for stochastic optimization,''
  \emph{arXiv preprint arXiv:1412.6980}, 2014.

\bibitem{salimbeni2018natural}
H.~Salimbeni, S.~Eleftheriadis, and J.~Hensman, ``Natural gradients in
  practice: {N}on-conjugate variational inference in {G}aussian process
  models,'' in \emph{International Conference on Artificial Intelligence and
  Statistics}, 2018, pp. 689--697.

\bibitem{klambauer2017self}
G.~Klambauer, T.~Unterthiner, A.~Mayr, and S.~Hochreiter, ``Self-normalizing
  neural networks,'' in \emph{Advances in Neural Information Processing
  Systems}, 2017, pp. 971--980.

\bibitem{kim2008outlier}
H.-C. Kim and Z.~Ghahramani, ``Outlier robust {G}aussian process
  classification,'' in \emph{Joint IAPR International Workshops on Statistical
  Techniques in Pattern Recognition (SPR) and Structural and Syntactic Pattern
  Recognition (SSPR)}.\hskip 1em plus 0.5em minus 0.4em\relax Springer, 2008,
  pp. 896--905.

\bibitem{ruiz2018augment}
F.~J. Ruiz, M.~K. Titsias, A.~B. Dieng, and D.~M. Blei, ``Augment and reduce:
  {S}tochastic inference for large categorical distributions,'' in
  \emph{International Conference on Machine Learning}, 2018, pp. 4400--4409.

\end{thebibliography}

\end{document}